\pdfoutput=1

\documentclass[11pt]{article}

\usepackage{acl}

\usepackage{times}
\usepackage{latexsym}

\usepackage[T1]{fontenc}

\usepackage[utf8]{inputenc}

\usepackage{microtype}

\usepackage{inconsolata}

\usepackage{multirow}
\usepackage{adjustbox}
\usepackage{booktabs} 
\usepackage{amsmath}  
\usepackage{graphicx}
\usepackage{subcaption}
\usepackage{caption}
\usepackage{amssymb}
\usepackage[shortlabels]{enumitem}
\usepackage{tcolorbox}
\usepackage{array}
\usepackage{pifont}
\usepackage{colortbl}
\usepackage{xcolor}
\title{Exploring the Impact of Table-to-Text Methods on Augmenting LLM-based Question Answering with Domain Hybrid Data}

\author{Dehai Min\thanks{~Equal Contributions.}$^{1,4}$ \quad  Nan Hu$^{*1,4}$  \quad Rihui Jin$^{1,4}$ \quad  Nuo Lin$^1$ \quad {\bf Jiaoyan Chen}$^2$ \\ \quad {\bf Yongrui Chen}$^{1,4}$\quad {\bf Yu Li}$^{1,4}$ \quad {\bf Guilin Qi}$^{1,4}$\thanks{~Corresponding author.}  \quad {\bf Yun Li}$^3$  \quad {\bf Nijun Li}$^3$ \quad {\bf Qianren Wang}$^3$  \\
      $^1$School of Computer Science and Engineering, Southeast University, China \\ $^2$Department of Computer Science, The University of Manchester, United Kingdom \\ $^3$Advanced Cognitive AI Lab, Huawei Technologies, China \\
      $^4$Key Laboratory of New Generation Artificial Intelligence Technology and Its\\ Interdisciplinary Applications (Southeast University), Ministry of Education, China \\
 \texttt{\{zhishanq, nanhu, gqi\}@seu.edu.cn}
      }

\begin{document}
\maketitle
\begin{abstract}
Augmenting Large Language Models (LLMs) for Question Answering (QA) with domain specific data has attracted wide attention. 
However, domain data often exists in a hybrid format, including text and semi-structured tables, posing challenges for the seamless integration of information. Table-to-Text Generation is a promising solution by facilitating the transformation of hybrid data into a uniformly text-formatted corpus. Although this technique has been widely studied by the NLP community, there is currently no comparative analysis on how corpora generated by different table-to-text methods affect the performance of QA systems.
In this paper, we address this research gap in two steps. First, we innovatively integrate table-to-text generation into the framework of enhancing LLM-based QA systems with domain hybrid data. Then, we utilize this framework in real-world industrial data to conduct extensive experiments on two types of QA systems (DSFT and RAG frameworks) with four representative methods: Markdown format, Template serialization, TPLM-based method, and LLM-based method. Based on the experimental results, we draw some empirical findings and explore the underlying reasons behind the success of some methods. We hope the findings of this work will provide a valuable reference for the academic and industrial communities in developing robust QA systems.

\end{abstract}

\section{Introduction}

Enhancing the performance of Large Language Models (LLMs) in domain-specific Question Answering (QA) has been a focus of research, predominantly employing two key approaches \cite{ling_domain_2023,wang_survey_2023-1}: Domain-Specific Fine-Tuning (DSFT) which involves training LLMs on the domain-specific corpus \cite{gururangan_dont_2020,wu_pmc-llama_2023}, and Retrieval-Augmented Generation (RAG) which utilizes a domain-specific corpus as an external knowledge base \cite{lewis_retrieval-augmented_2020}. These approaches, leveraging the inherent text processing strengths of LLMs, have been widely adopted in text-only scenarios, yielding significant improvements \cite{zhao2023domain}.

However, real-world data in many domains typically exists in a hybrid format, comprising not only text but also substantial volumes of semi-structured tables, as observed in e.g., scientific literature and medical reports \cite{chen2020hybridqa,zhu2021tat}. 
These tables frequently appear alongside text within the same document, providing semantically supplementary or complementary information crucial for a comprehensive understanding of the content \cite{chen2020open}.
In exploring the potential of leveraging hybrid data to enhance the performance of LLMs, it is crucial to effectively integrate these data, ensuring the coexistence of text and tables. The current methods for handling the heterogeneity of text and tables have significant drawbacks: 1) Directly flattening tables by concatenating cells row by row not only results in the loss of structural information embedded in the original table but also severs the informational links between cells \cite{sui_gpt4table_2023,xie_unifiedskg_2022}. 2) Mapping text and tables to different vector spaces separately and then integrating them, not only increases complexity but also disrupts the semantic connection between the two types of data \cite{li2021dual,huang_mixed-modality_2022}.

One promising solution is table-to-text generation \cite{luo_few-shot_2022,cheng-etal-2022-hitab}, which aims to generate natural language statements that faithfully describe the information in the provided table. Through this, we can transform hybrid data into a unified natural language representation that is more suitable for use by LLMs, while also preserving the important information from the tables and the semantic connections between the data.
Although table-to-text generation has been widely studied by the NLP community, there is currently no comparative analysis on how corpora generated by different table-to-text methods affect the performance of domain-specific QA systems.

In this work, we address this research gap by two steps. First, we innovatively integrate table-to-text generation into the framework of enhancing LLM-based QA systems with domain hybrid data. Then, we utilize this framework to conduct extensive experiments on two types of QA systems (DSFT and RAG paradigms) with four representative table-to-text methods. We choose the following four strategies: 1) \textbf{Markdown} format; 2) \textbf{Template} serialization; 3) \textbf{TPLM-based} method; 4) \textbf{LLM-based} method.
These strategies differ in complexity and underlying technology. The Markdown and Template serialization offer simplicity, while the TPLM-based and LLM-based methods leverage the capabilities of advanced language models to generate more nuanced text. 

In terms of implementation, we collect a real-world hybrid dataset called ICT-DATA, by extracting text and tables from numerous documents about Information and Communication Technology (ICT) products. It is important to note that the text contained in tables accounts for approximately 18\% of the total content in ICT-DATA (based on word count statistics).
We employ different table-to-text methods to process the tables in ICT-DATA, obtaining different ICT corpora. These corpora are then utilized to build QA systems. Moreover, we create a benchmark dataset named ICTQA, which consists of QA pairs based on the knowledge of ICT-DATA. This dataset is particularly suitable for evaluating enhanced LLMs, as it includes some industry-specific knowledge not covered in the general LLMs training stage.

To our knowledge, our research is the first to comprehensively compare different table-to-text strategies on LLM-based QA systems enhanced by domain hybrid data. Our main findings are as follows:
\begin{itemize}[topsep=0pt, itemsep=0pt, leftmargin=.2in, parsep=0pt]
\item Table-to-text methods significantly impact the performance of QA systems, with relative score differences ranging from 2.8\% to 9.0\% in human evaluation and 4.8\% to 16\% in GPT-4 evaluation. In two systems, selecting the appropriate method can yield considerable benefits.
\item In the DSFT paradigm, LLM-based and TPLM-based consistently outperform others across various model settings, demonstrating their superiority. In the RAG paradigm, while the LLM-based method still performs excellently, the Markdown has shown unexpected effectiveness.
\item The varying frequency of domain-specific terms and verbs produced by these methods, alongside the differing quality of semantic representations in the generated text chunks, which appear to be pivotal factors influencing performance disparities across the two systems.
\end{itemize}

\begin{figure*}[ht!]
\centering
     \includegraphics[width=0.98\textwidth]{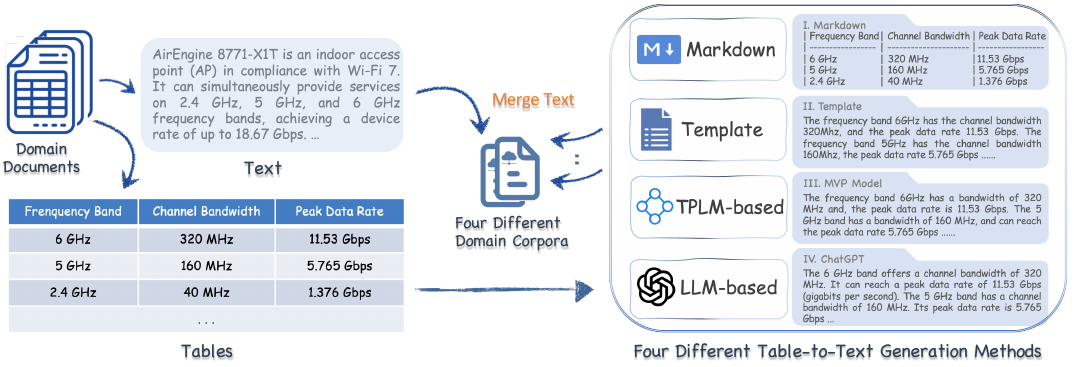}
     \caption{Illustration of four domain corpora generation process. Different table-to-text methods are applied to tables of domain documents, generating different text. These generated texts are then merged with the original document texts, yielding different domain corpora.}
     \label{fig:table_to_text_framework}
     \vspace{-4mm}
\end{figure*}

\section{Table-to-Text Generation}
\label{sec:table_to_text_gene}

Table-to-text generation \cite{parikh2020totto, chen_logical_2020, cheng-etal-2022-hitab} aims to create natural language descriptions from semi-structured tabular data, such as web tables. As shown in Figure~\ref{fig:table_to_text_framework}, we apply four representative table-to-text methods to textualize the tables in ICT-DATA, forming four different corpora. Formally: Let \( F_i: \text{Table} \rightarrow \text{Text} \) represent four table-to-text functions for \( i = 1, 2, 3, 4 \). With the original ICT-DATA \( D = \{\text{Tab}, \text{Text}\} \), each \( F_i \) converts tables \( \text{Tab} \) into text. The resulting ICT Corpora \( C_i \) are formed by combining these texts with \( \text{Text} \):
\[ C_i = F_i(\text{Tab}) \cup \text{Text}, \quad i = 1, 2, 3, 4 \]
We next provide a detailed introduction of these four methods. Table \ref{tab:table_to_text_dif} provides a comparative analysis of these methods in terms of their resource requirements, processing speeds, and text diversity.

\begin{table}[b]
\vspace{-2mm}
\centering\resizebox{0.47\textwidth}{!}
{
\begin{tabular}{l|cccc}
\toprule
\textbf{Method}   & \textbf{Resource} & \textbf{Speed}  & \textbf{Diversity}\\ \hline
\textbf{Markdown}       & CPU           & Fast      & Low                 \\
\textbf{Template}       & CPU           & Fast      & Moderate       \\
\textbf{TPLM-based}      & GPU           & Moderate  & High              \\
\textbf{LLM-based}      & GPU or API    & Low       & Very High  \\
\bottomrule
\end{tabular} 
} 
\vspace{-2mm}
\caption{Comparison of table-to-text methods: resource usage, generation speed and diversity of generated text.}
\label{tab:table_to_text_dif}
\vspace{-4mm}
\end{table}

\begin{itemize}[topsep=0pt, itemsep=0pt, leftmargin=.1in, parsep=0pt]
\item \textbf{Markdown} format: A straightforward method to represent tables in Markdown format. It does not involve model training and can be rapidly processed via scripts without manual intervention.
\item \textbf{Template} serialization: This method uses a set of templates designed based on table features for textualization \cite{li2023table, ye2019variational}. It achieves slightly higher diversity in the generated text compared to the Markdown method, attributed to the use of multiple pre-prepared templates to accommodate different types of tables, which requires some manual involvement.
\item \textbf{TPLM-based} method: This method involves fine-tuning Traditional Pre-trained Language Models (TPLMs), such as T5 \cite{raffel2020exploring} and BART \cite{lewis-etal-2020-bart}, on specific table-to-text generation task datasets \cite{liu-etal-2022-plog}. In this paper, we utilize the MVP model \cite{tang_mvp_2023}, which initially pre-trains the BART model on numerous natural language generation datasets, followed by fine-tuning on various cross-domain table-to-text datasets. It allows customized adjustment of the output through fine-tuning, offering higher flexibility and domain adaptability, while requiring more computational resources.
\item \textbf{LLM-based} method: Recent endeavors employing LLMs for this task have drawn significant attention \cite{bian2023hellama}. Impressively, \citet{zhao-etal-2023-investigating} demonstrate that GPT-* models often outperform the best-performing fine-tuned models. We refer to their findings and utilize ChatGPT in a one-shot setting in our work. Similar to TPLM-based methods, this approach can be custom-tailored using In-Context Learning. Moreover, using the APIs of certain proprietary LLMs might pose risks of domain data leakage.
\end{itemize}
Some examples of table-to-text, along with the specific templates and prompts for ChatGPT used in this paper, can be found in Appendix~\ref{sec:appendix_table_to_text_details_examples}.


\begin{figure}[ht]
    \centering
    \begin{subfigure}[b]{0.48\textwidth}
        \centering
        \includegraphics[width=\textwidth]{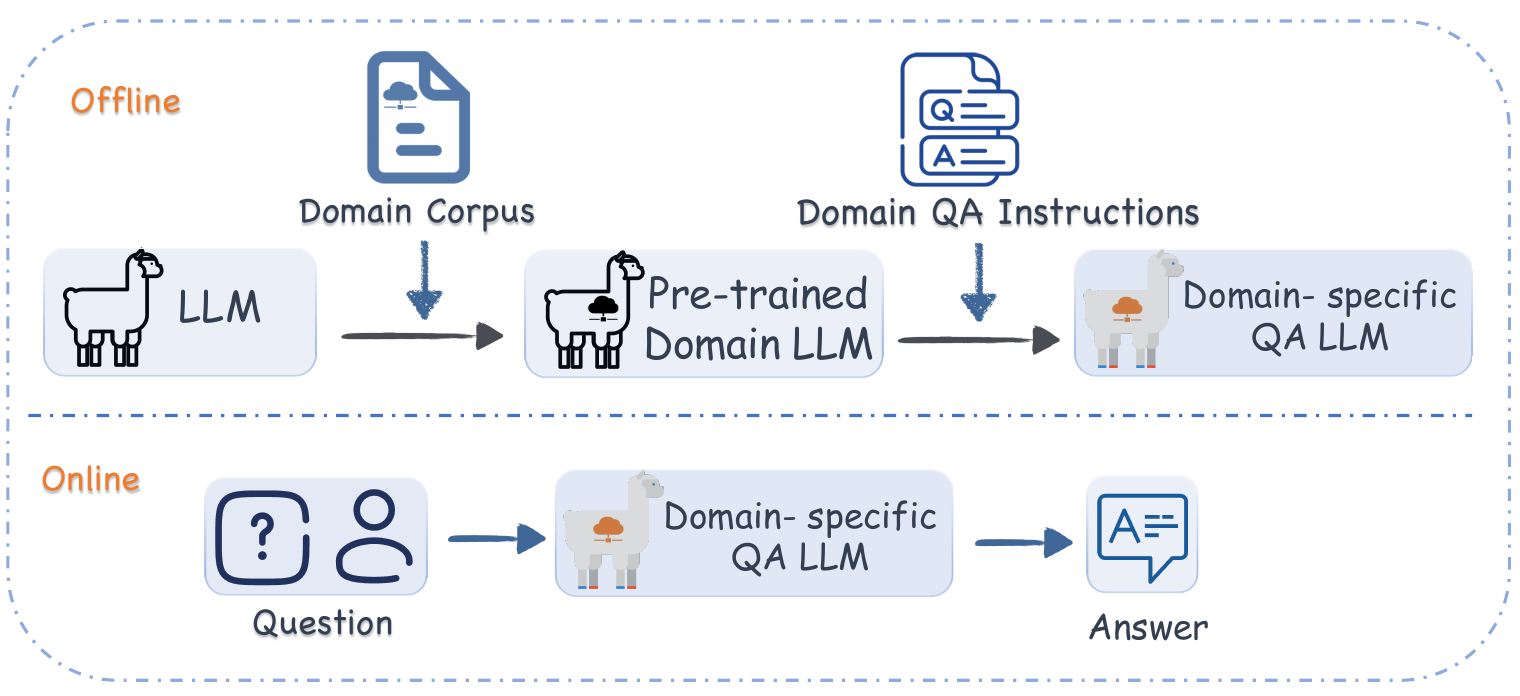}
        \caption{Domain-Specific Fine-Tuning QA system}
        \label{fig:QA_framework_dsft}
    \end{subfigure}
    
    \begin{subfigure}[b]{0.48\textwidth}
        \centering
        \includegraphics[width=\textwidth]{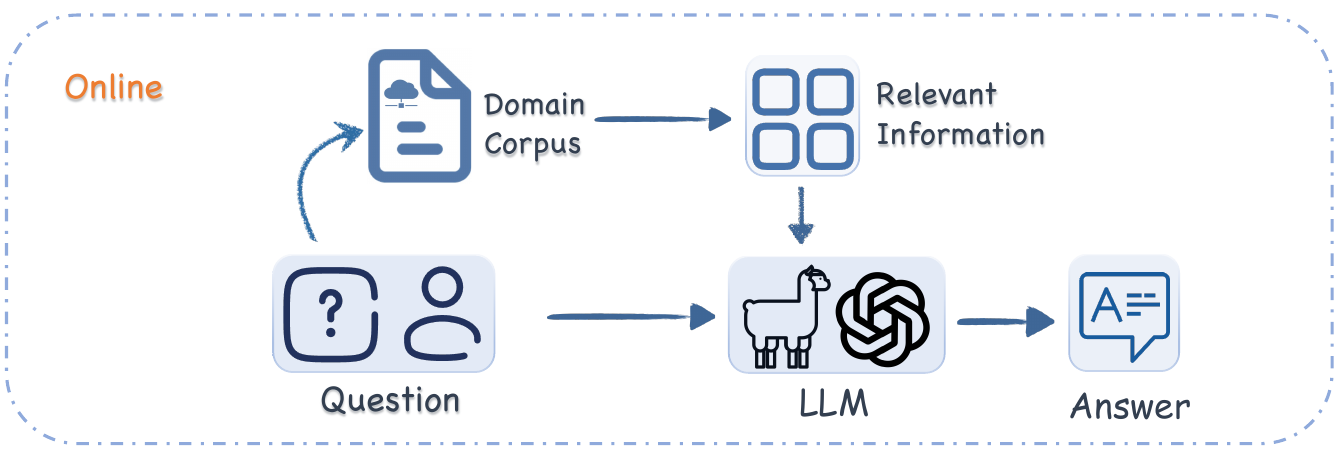}
        \caption{Retrieval-Augmented Generation QA system}
        \label{fig:QA_framework_rag}
    \end{subfigure}
    \caption{Framework of domain-enhanced QA systems.}
    \label{fig:QA_framework}
    \vspace{-3mm}
\end{figure}

\section{Building LLM-based QA Systems with Domain Corpora} 
We will introduce separately how two LLM-based QA systems utilize these corpora. Their framework overview can be viewed in Figure~\ref{fig:QA_framework}.

\noindent\textbf{Domain-Specific Fine-Tuning.} 
In this approach, we first pre-train the LLM on the ICT corpus using next-token prediction \cite{radford2018improving}, enabling the model to incrementally learn domain knowledge. Subsequently, we adapt the model to the QA task through instruction tuning \cite{ouyang2022training}. Formally, an original LLM \( M \), is pre-trained on each ICT Corpus \( C_i \), to obtain an updated foundation model \( M_i' \):
   \[ M_i' = \text{Pre-Train}(M, C_i), \quad i = 1, 2, 3, 4 \]
The updated models are then further trained on the same instruction set \( I \) tailored for the QA task, resulting in the final QA oriented models \( M_i^{QA} \):
   \[ M_i^{QA} = \text{FineTune}(M_i', I), \quad i = 1, 2, 3, 4 \]

\noindent\textbf{Retrieval-Augmented Generation.}  
In this paradigm, we adopt the framework proposed by LangChain \cite{LangChain} with the Dense Passage Retriever (DPR) method \cite{karpukhin_dense_2020}, which consists of a multi-step process: 1) Splitting the large-sized Corpus \( C_i \) into smaller chunks \(\{p_j\}^{C_i} \); 2) Encoding each text chunk \( p_j \) into a d-dimensional vector by an encoder \( E_{P}(\cdot) \), which captures its semantic essence; 3) Building an indexed Vector Store for these vectors, optimizing the storage for efficient retrieval; 4) For each query \( Q \), retrieving the \(K\) most relevant text chunks, \({P=}\{p_k\}_{k=1}^K \); 5) Using both the query \( Q \) and the retrieved prompts \( P \) to generate the final answer with the LLM.

\section{Dataset and Evaluation Metrics}

\subsection{Evaluation Dataset}

\noindent\textbf{ICT-DATA.}
We collect ICT-DATA based on 170 English technical documents related to ICT products. Each product document consists of tables and text, whose contents include product descriptions, configuration guides, terms, and definitions, etc. The total storage size is approximately 6GB. Moreover, the number of words in the table data accounts for about 18\% of the total number of words in the dataset. In Appendix~\ref{sec:appendix_ictdata_dataset}, we provide detailed statistics and the preprocessing methods used for the table data.

\noindent\textbf{ICTQA.}
We create the ICTQA dataset to evaluate the performance of domain QA systems, by collecting 9,000 questions with long-form answers from the actual ICT product technical support QA platform. All the answers are written by experts based on product documents. We manually select 500 questions as the test set, whose answers involve knowledge from both tables and text. The remaining QA pairs are used as the training set for the instruction fine-tuning phase in the DSFT paradigm. We show statistics and some examples in Appendix~\ref{sec:appendix_ictqa_dataset}.

\subsection{Evaluation Metrics}
To evaluate the model's responses, we employ both automated and manual evaluation methods. 

\noindent\textbf{Automated Evaluation Metrics.} 
Given that traditional lexical-overlap-based metrics (such as BLEU and ROUGE) are inadequate for evaluating the quality of long-form answers generated by LLMs \cite{krishna_hurdles_2021, kamalloo-etal-2023-evaluating}, we use GPT-4 as an evaluator with a demonstration setting, scoring responses based on their similarity to the golden answer \cite{liu-etal-2023-g}.
The score ranges from 0 to 5 with discrete values; 0 indicates incoherent answers with repeated fields or responses like ``I don't know the answer'', 1 represents minimal similarity to the golden answer, and 5 denotes an accurate answer.

\noindent\textbf{Human Evaluation.} 
Given the limitations in evaluating long-form answers using existing automated metrics \cite{wang-etal-2023-chatgpt,kamalloo-etal-2023-evaluating}, three evaluators with domain knowledge are asked to score responses based on the helpfulness and similarity to the golden answer, using the same scoring criteria with a range of 0 to 5 as the GPT-4 evaluator. 

For fairness and to eliminate potential bias, responses are presented anonymously to both the GPT-4 and human evaluators. The full prompt, evaluation setup for human and scoring criteria are detailed in Appendix~\ref{sec:appendix_evaluation_setup}.

\section{Experimental Setup}

\noindent\textbf{QA Systems of the DSFT Paradigm.}
Within the DSFT paradigm, we utilize Meta's OPT (1.3B to 13B) \cite{zhang_opt_2022} and Llama2-base (7B, 13B) \cite{touvron2023llama} as foundation models. The OPT models offer variable sizes to enhance robustness. To mitigate training costs, we employ the QLoRA \cite{dettmers_qlora_2023} strategy for pre-training and instruction fine-tuning. The instruction template can be found in Appendix~\ref{sec:appendix_instruction_template}.

\noindent\textbf{QA Systems of the RAG Paradigm.}
We use the Llama2-chat models (7B, 13B, and 70B) and GPT-3.5-turbo for inference. We divide the corpus into smaller chunks, ensuring the integrity of sentences and keeping their lengths below 3000 characters. Subsequently, text chunks are vectorized using the BGE embedding model \cite{zhang_retrieve_2023}. We utilize the FAISS library \cite{8733051} to retrieve the vectors of the top-3 relevant text chunks based on similarity. These chunks are input to the LLM with the corresponding questions for answering through the RAG-Chain from LangChain \cite{LangChain}.

\noindent\textbf{Fair Comparison.} 
To maintain consistency and control variables, all models are trained or used under the same settings on four different corpora. Detailed training parameters and GPU costs are available in Appendix~\ref{sec:appendix_training_setup}.

\begin{table*}[!ht]
    \centering
    \begin{adjustbox}
     {max width=0.99\textwidth}
    \begin{tabular}{c|c|cccccc|cccc}
    \hline
     \multirow{2}{*}{\textbf{Metrics}} & \multirow{1}{*}{\textbf{Table-to-Text }} & \multicolumn{6}{c|}{\textbf{Domain-Specific Fine-Tuning}} & \multicolumn{4}{c}{\textbf{Retrieval-Augmented Generation}} \\
     
    & \textbf{Method} & \textbf{OPT-1.3B} & \textbf{OPT-2.7B} & \textbf{OPT-6.7B} & \textbf{OPT-13B} & \textbf{Llama2-7B} & \textbf{Llama2-13B} & \textbf{GPT-3.5-turbo} & \textbf{Llama2-7B} & \textbf{Llama2-13B} & \textbf{Llama2-70B} \\

    \hline
\multirow{4}{*}{\textbf{Human}} & \textbf{Markdown}      & 2.05 & 2.41 & 2.38 & 2.51 & 2.82 & 3.05    & 3.29 & \textbf{3.72} & \underline{3.98} & \underline{3.94} \\

& \textbf{Template}      & 2.04 & 2.40 & 2.26 & 2.47 & 2.82 & 3.04    & \underline{3.36} & 3.44 & 3.96 & 3.76 \\

\multirow{2}{*}{\textbf{Eval.}} & \textbf{TPLM-based}     & \underline{2.12} & \underline{2.43} & \underline{2.43} & \underline{2.58} & \textbf{3.20} & \underline{3.13}    & 3.26 & 3.27 & 3.92 & 3.64 \\

& \textbf{LLM-based} & \textbf{2.18} & \textbf{2.57} & \textbf{2.51} & \textbf{2.62} & \underline{2.96} &  \textbf{3.19}    & \textbf{3.62} & \underline{3.71} & \textbf{4.26} & \textbf{4.09}  \\

\cline{2-12}
& \textbf{RSD(\%)} & 2.80 & 3.40 & 5.00 & 3.00 & 7.60 & 3.00 & 7.20 & 9.00 & 6.80 & 9.00 \\

\hline
    \addlinespace
\hline

\multirow{4}{*}{\textbf{GPT-4}} & \textbf{Markdown}     & 1.74 & 2.16 & 2.27 & 2.25 & 2.7 & 3.06    & \underline{3.28} & \textbf{3.66} & \underline{3.67} & \textbf{3.74} \\

& \textbf{Template}      & 1.81 & 2.22 & 2.39 & 2.34 & 2.84 & 3.08    & 3.27 & 3.06 & 3.38 & 3.37 \\

\multirow{2}{*}{\textbf{Eval.}}& \textbf{TPLM-based}     & \underline{2.33} & \underline{2.46} & \underline{2.45} & \underline{2.53} & \textbf{3.20} & \underline{3.19}    & \underline{3.28} & 2.9 & 3.41 & 3.30 \\

& \textbf{LLM-based} & \textbf{2.57} & \textbf{2.69} & \textbf{2.73} & \textbf{2.86} & \underline{3.06} &  \textbf{3.30}    & \textbf{3.64} & \underline{3.59} & \textbf{3.69} & \underline{3.54}  \\

\cline{2-12}
& \textbf{RSD(\%)} & 16.60 & 10.60 & 9.20 & 12.20 & 10.00 & 4.80 & 7.40 & 15.20 & 6.20 & 8.80  \\

\hline

\end{tabular}
\end{adjustbox}
\caption{
The average scores from Human Evaluation and GPT-4 Evaluation of the QA systems with four representative table-to-text methods. In each setting, the best result is shown in bold, and the second-best result is underlined. Relative Score Difference (RSD) is calculated using the formula \((\text{Highest Score} - \text{Lowest Score}) / 5\).
}
\label{tab:all_results_human_and_gpt4}

\end{table*}

\section{Results}
In the following subsections, we will discuss three research questions regarding our study.

\subsection{RQ1: How do these methods affect the performance of QA systems?}

\begin{figure}[t]
\centering 
     \includegraphics[width=0.43\textwidth]{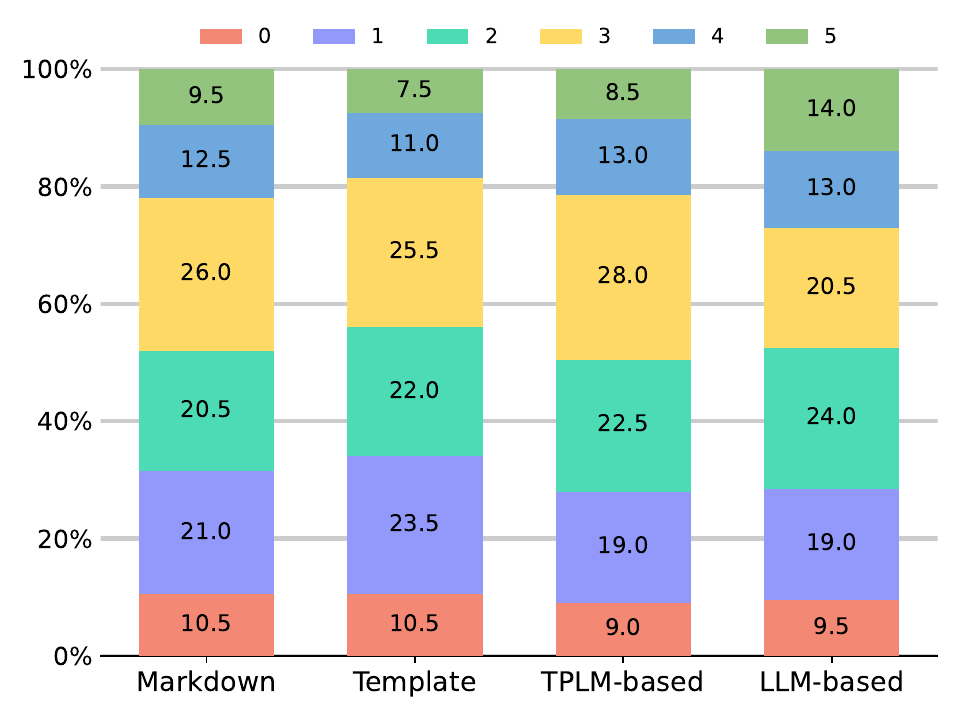}
     \caption{The scores distribution from human evaluation for the DSFT QA systems based on OPT-6.7B.}
     \label{fig:6p7B_score_distribution}
\end{figure}

\begin{figure*}[ht]
    \centering
    \begin{subfigure}[b]{0.33 \textwidth}
        \centering
        \includegraphics[width=\textwidth]{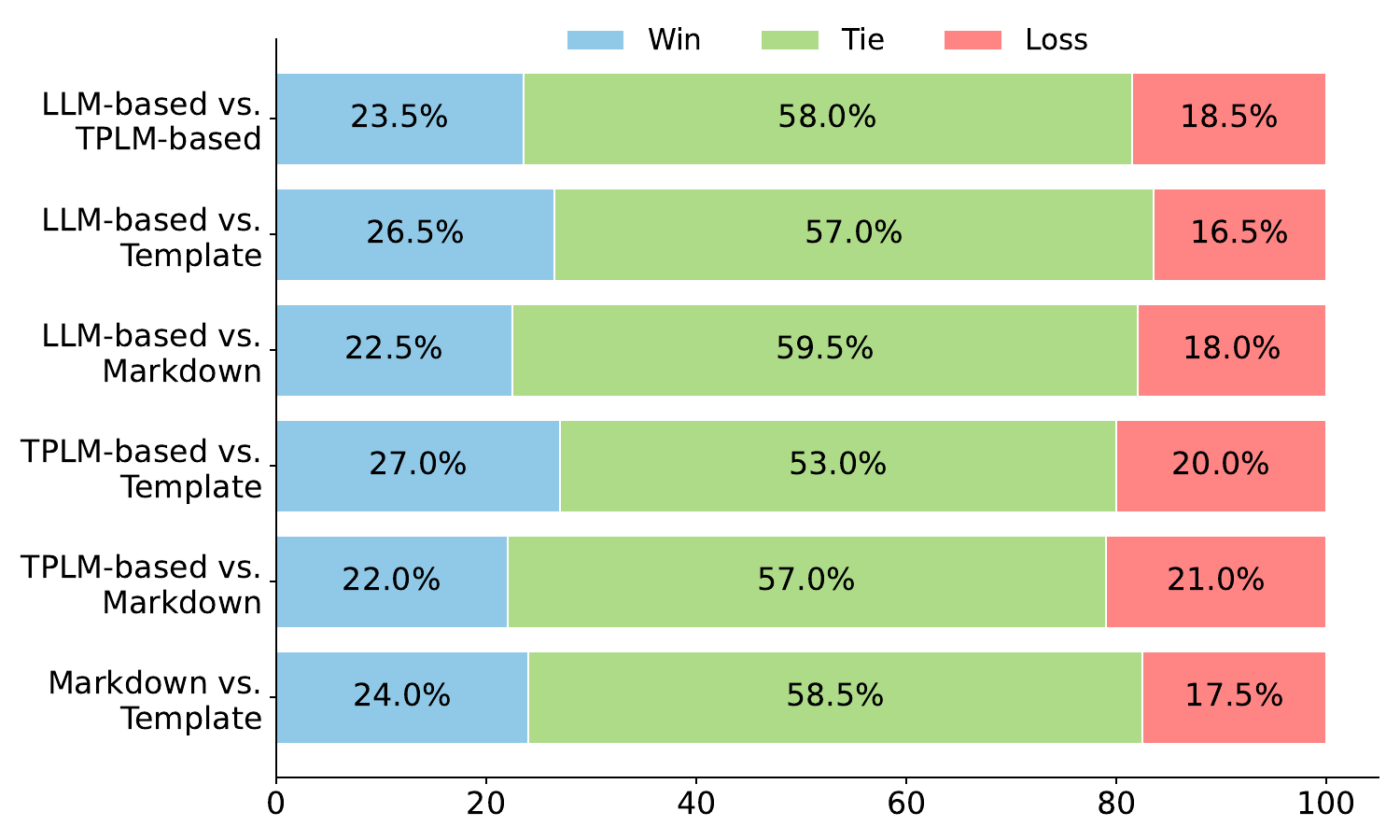}
        \caption{OPT-6.7B in DSFT Paradigm}
        \label{fig:OPT-6p7B_dsft}
    \end{subfigure}%
    \hfill
    \begin{subfigure}[b]{0.33\textwidth}
        \centering
        \includegraphics[width=\textwidth]{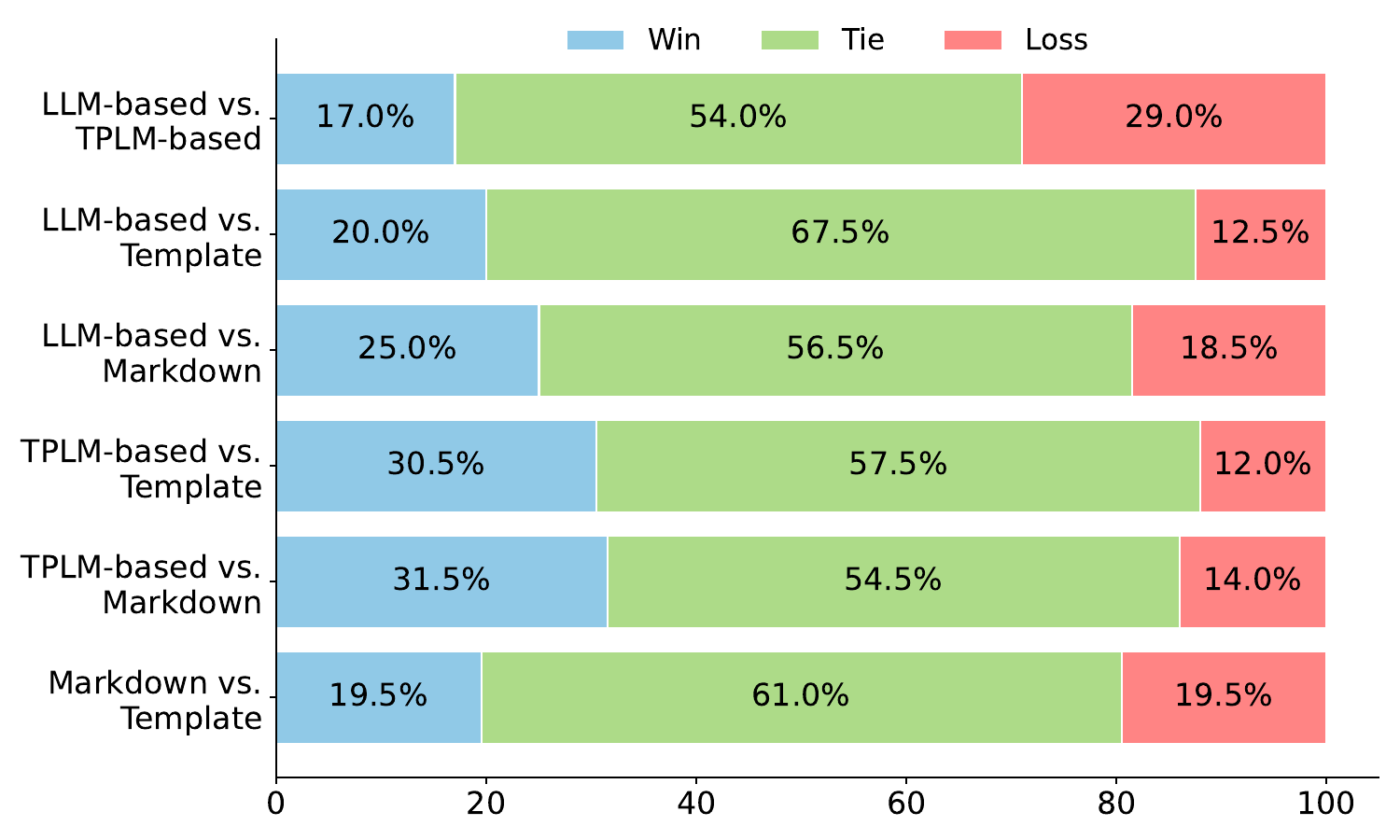}
        \caption{Llama2-7B in DSFT Paradigm}
        \label{fig:Llama2-7B_dsft}
    \end{subfigure}%
    \hfill
    \begin{subfigure}[b]{0.33\textwidth}
         \centering
         \includegraphics[width=\textwidth]{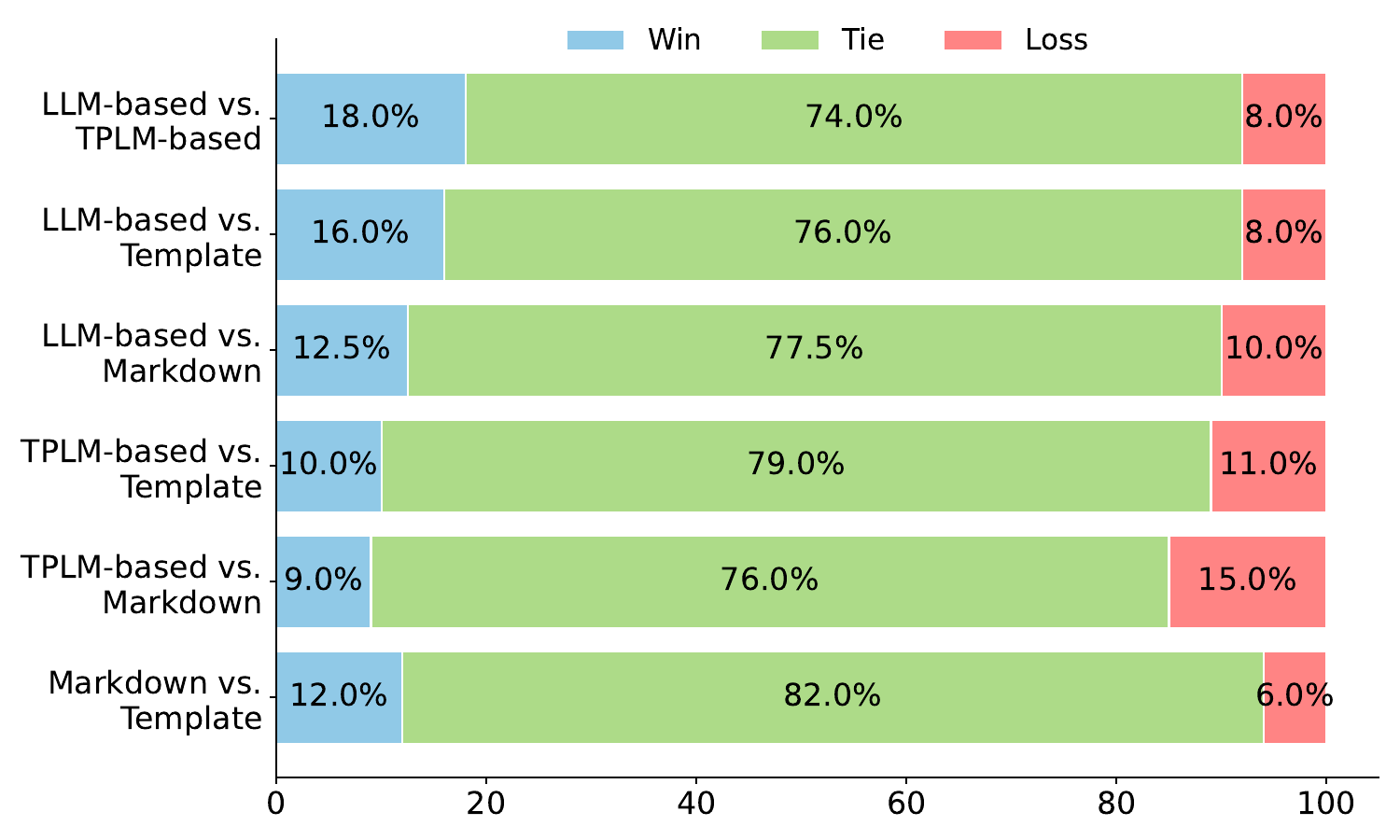}
         \caption{Llama2-70B in RAG Paradigm}
         \label{fig:rag_gpt3p5}
    \end{subfigure}
    \caption{Comparison of human evaluation scores between QA models using different Table-to-Text methods. `A vs. B win' indicates the percentage of test set instances where Model A's score surpasses Model B's.}
    \label{fig:win_rate_overall}
    \vspace{-4mm}
\end{figure*}

Table~\ref{tab:all_results_human_and_gpt4} shows the average scores for different QA system setups on the ICTQA test set. We can see that there are significant differences in the performance of the two types of QA systems enhanced by corpora generated from different table-to-text methods. Their Relative Score Differences range from 2.8\% to 9.0\% in human evaluation and from 4.8\% to 16\% in GPT4 evaluation. For a more detailed observation, we present the score distribution from human evaluation of the DSFT QA models based on OPT-6.7B in Figure~\ref{fig:6p7B_score_distribution}. From this figure, we can observe significant differences in score distribution among different QA models, reflecting their performance variations. From Table \ref{tab:all_results_human_and_gpt4}, we note that in the DSFT paradigm, both TPLM-based and LLM-based methods, which utilize language models for table-to-text generation, perform well across different models. Particularly, the LLM-based method shows the best performance in many models. On the other hand, the RAG paradigm provides a different observation. While the LLM-based method continues to exhibit excellent performance, the Markdown format shows a significant and unexpected improved performance in the RAG paradigm compared to DSFT, even best-performing in some models. To further illustrate these findings, we show the competition results of some QA system scores in Figure~\ref{fig:win_rate_overall}. We can clearly observe that the methods with higher average scores also have a higher probability of achieving better scores for each question. These observations underscore the necessity of choosing the appropriate method for processing table data when building domain-specific QA systems.

\subsection{RQ2: What are the potential reasons for their different performances?}
\label{sec:reason_for_markdown_work_on_rag}

Since DSFT and RAG systems utilize domain corpora in different ways, we will discuss them separately in this section.

\begin{table}[ht]
\centering\resizebox{0.49\textwidth}{!}{
\begin{tabular}{l|cccc}
\toprule

\textbf{Freq (k)}   & \textbf{\(C_{1}\cdot\) Markdown} & \textbf{\(C_{2}\cdot\) Template} & \textbf{\(C_{3}\cdot\) TPLM-based} & \textbf{\(C_{4}\cdot\) LLM-based} \\ \hline
\textbf{Term}                & 821 & 1040      & 2358  & 2254 \\ 
\textbf{Verbs }                 & 313   & 315      & 682      & 1207    \\
\bottomrule
\end{tabular}
}
\vspace{-2mm}
\caption{Absolute frequency of verbs and terms contained in the corpora \( C_i \) generated by different methods.}
\label{tab:dif_between_content_method}
\vspace{-2mm}
\end{table}

\noindent\textbf{For the DSFT paradigm.}  Inspired by the findings of \cite{pmlr-v202-biderman23a, razeghi-etal-2022-impact, elazar_measuring_2023}, which suggest a correlation and causal relationship between the ability of LLMs to answer factual questions and the frequency of salient entities found in their pre-training corpora, we also observe that different table-to-text methods have inconsistent preferences for domain verbs when describing  tables. Following the approach of \cite{zevallos_frequency_2023, wang-etal-2023-self-instruct}, we extract domain term sets and related verb sets from the QA pairs in the ICTQA test set. We then calculate the absolute frequency of these terms and verbs as they appear in the corpora generated by different table-to-text methods. In Table~\ref{tab:dif_between_content_method}, we can clearly see significant differences in these frequencies across different corpora. For example, LLM-based methods show a term frequency more than twice that of Template methods, with verb frequency quadrupling. This is because LLM-based methods tend to supplement the subject with the domain entity corresponding to the attribute when describing tables, and exhibits greater diversity in verbs. In contrast, Template methods use more pronouns, such as `it', and monotonous predicates (usually `be' verbs).
By comparing these frequency rankings with the performance shown in Table~\ref{tab:all_results_human_and_gpt4}, we can observe a positive correlation between them: methods with higher frequencies, especially the TPLM and LLM-based methods, correspond to superior QA capabilities in the DSFT systems.

\begin{figure}[t]
\centering 
     \includegraphics[width=0.42\textwidth]{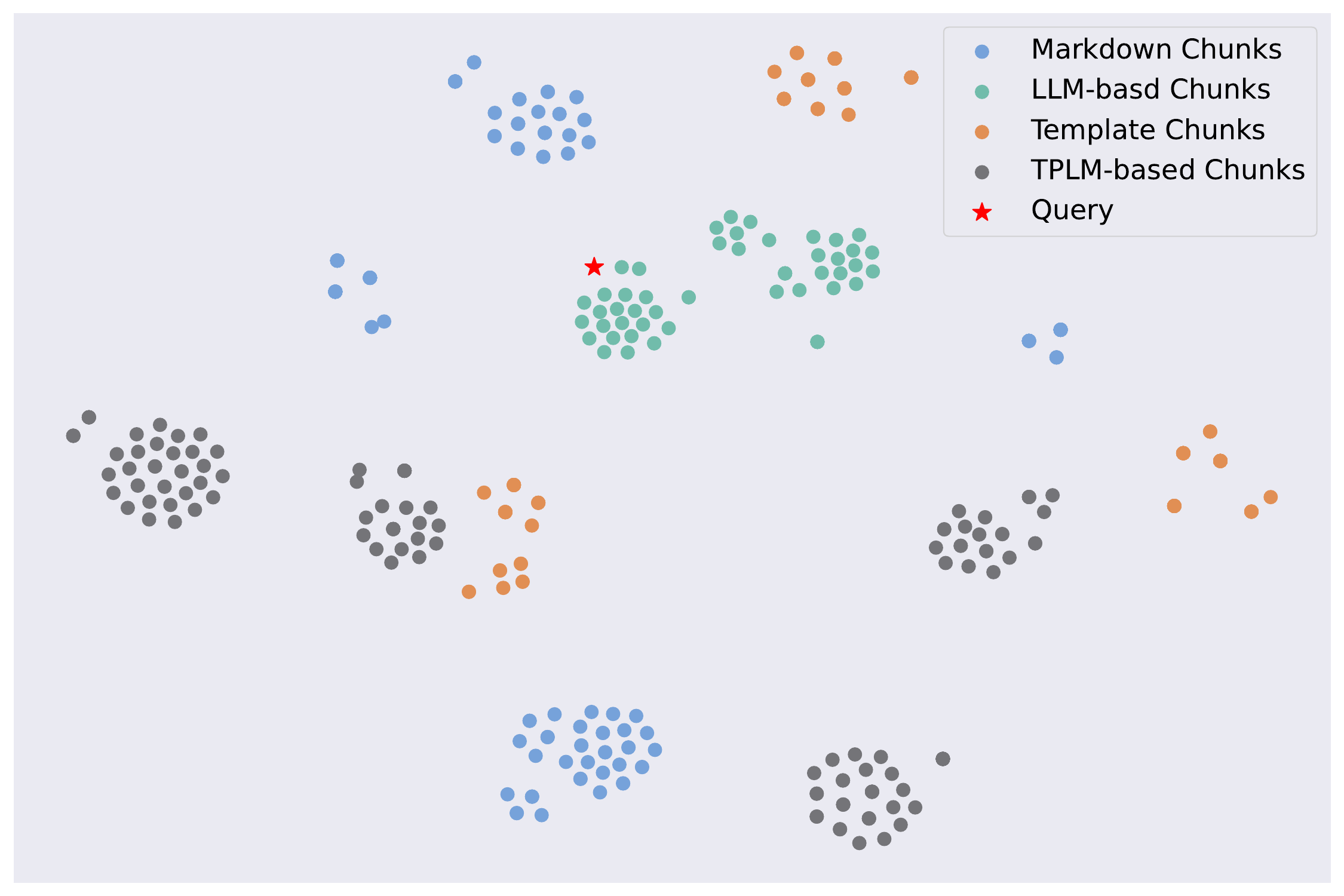}
     \caption{A t-SNE visualization of chunk clusters in the embedding space of the RAG system. `X Chunks' represents chunks related to the query (red star) from the corpus generated by X table-to-text method.}
     \label{fig:embedding_vis_four_method}
     \vspace{-4mm}
\end{figure}

\noindent\textbf{For the RAG paradigm.} Under the same LLM reader setup, retrieval accuracy in this semantic space crucially impacts RAG performance \cite{ma-etal-2023-query}. The retrieval process involves selecting the vectorized chunks with the highest similarity scores to the query vector. To investigate the impact of different methods on retrieval effectiveness, we use t-SNE \cite{van2008visualizing} to visualize the clustering of a query and related chunks in the semantic space at Figure~\ref{fig:embedding_vis_four_method}. It could be clearly seen that chunks generated by the LLM-based and Markdown methods, which perform well in Table~\ref{tab:all_results_human_and_gpt4}, are closer to the query in the semantic space. This makes the chunks related to the query more likely to be retrieved, thereby improving the system's performance. This suggests that in the RAG framework with the DPR method, the texts generated by these methods have more retrieval-friendly semantic representations and better alignment between queries and documents.

\begin{table}[h]
\centering\resizebox{0.49\textwidth}{!}{
\begin{tabular}{l|cccc}
\toprule

\textbf{Freq (Avg.)}   & \textbf{Markdown} & \textbf{Template} & \textbf{TPLM-based} & \textbf{LLM-based} \\ \hline
\textbf{Text Len} & 998 & 1259      & 1138  & 897 \\ 
\bottomrule
\end{tabular}
}
\vspace{-2mm}
\caption{The average length of text generated by different methods for each table.}
\label{tab:dif_avg_length_for_rag}
\vspace{-3mm}
\end{table}

\subsection{RQ3: Are there practical suggestions for choosing table-to-text methods?}
Through the analysis of RQ1 and RQ2, we know that the LLM-based strategy with ChatGPT is outstanding and reliable in both frameworks. In case its drawbacks mentioned in Section~\ref{sec:table_to_text_gene} are unacceptable, the TPLM-based strategy (i.e., selecting a well-tuned table-to-text model) is a good alternative in the DSFT paradigm. In the RAG paradigm, the simple and easy-to-use Markdown strategy is also a viable substitute. Additionally, although RAG systems using these four methods significantly outperform DSFT systems in terms of performance, building a vector retrieval library demands substantial memory resources. Therefore, referring to Table~\ref{tab:dif_avg_length_for_rag}, choosing methods that generate more concise texts, such as LLM-based and Markdown strategies, is a wise decision.

\subsection{Additional discussion on experimental results}
As shown in Table~\ref{tab:all_results_human_and_gpt4}, under the ICT dataset and the experimental setup of this study, the RAG method outperforms the DSFT method in Llama2 models. This demonstrates that RAG has an excellent performance as a lower cost method. We attribute this result to two main reasons: 1). The ICT data used in this study covers dense domain knowledge, and it is still challenging to adapt the LLM well to this complex domain data through incremental pre-training. 2). As the statistical analysis in Appendix~\ref{sec:appendix_ictqa_dataset}, most of the questions in the ICTQA are quizzes on the knowledge of product manuals. In this scenario, the existing excellent dense vector retrievers have high recall accuracy. The studies of \cite{gupta2024rag} and \cite{soudani2024fine} have respectively conducted detailed experiments on the choice between Fine-Tuning and RAG under the agricultural domain data and Less Popular Knowledge scenarios. Our experimental results in this work further validate their viewpoints. It is also worth noting that in this study, the bge-large-en embedding model \cite{zhang_retrieve_2023} embeds text chunks into 1024-dimensional vectors. During the retrieval of relevant chunks based on the questions, the peak running memory requirement is approximately 280G.

Another interesting experimental result is that GPT-3.5-turbo performs worse than the Llama2 family in the RAG paradigm. We manually observe the QA cases and find that GPT-3.5-turbo has a significantly higher probability of outputting ``I don't know the answer.'', even if the retriever finds text chunks containing the correct answer.

\section{Related Work}

\subsection{Domain Augmented Large Language Models.} 
In order to enhance the capabilities of LLMs in domain-specific tasks, some works develop LLMs through incremental training on an extensive domain corpus, inheriting the benefits of both the emergent abilities of LLMs and domain-specific knowledge \cite{luo_biomedgpt_2023,huang2023acegpt}. This technology yields significant results, but it demands substantial computational resources and incurs high costs \cite{wang_survey_2023-1}. In order to overcome this difficulty, a prompt-based solution that does not require updating model parameters has been proposed. They retrieve relevant domain information from external knowledge bases before answering questions with LLMs \cite{gao2023retrievalaugmented,wang2023augmenting,xu-etal-2023-retrieval}.

\subsection{Question Answering over Hybrid Data}
Some works study QA tasks on hybrid data that contain both tables and text \cite{zhu2021tat, chen2020hybridqa, chen2020open}. Popular approaches often involve designing a complex system that has independent modules to process text and tables separately. The information from these two modules is then merged and fed into a language model to generate answers \cite{zhong2022reasoning}. Additionally, some of these methods not only require annotations of metadata identifying text and tables relevant to the question, but they also rely on the formulation of executable languages to access tables, such as SQL or SPARQL \cite{nan_fetaqa_2022,li2021dual}. These executable languages often have strict assumptions about the structure of the tables. These limitations make these approaches ill-suited for the real-world LLM-based scenario domain QA systems. Therefore, the results of this study were not compared with these baseline models in the experiments.

\section{Conclusion}
This paper studies the impact of different table-to-text methods on LLM-based QA systems enhanced by domain hybrid data. Specifically, we meticulously compared four representative methods: Markdown formatting, Template serialization, TPLM-based, and LLM-based approaches. Through experiments, we show the superiority of the LLM-based and TPLM-based methods in the DSFT framework, and the excellence of the LLM-based and Markdown methods in the RAG framework. A key discovery is the varying frequency of domain-specific terms and verbs produced by these methods, alongside the differing quality of semantic representations in the generated text chunks, which appear to be pivotal factors influencing performance disparities across the two systems. These insights not only shed light on the nuances of table-to-text generation methods but also have profound implications for the enhancement of LLMs. Furthermore, they offer practical guidance for tailoring domain-specific QA systems to meet particular needs.

\section*{Acknowledgements}
This work is partially supported by National Nature Science Foundation of China under No. U21A20488. We thank the Big Data Computing Center of Southeast University for providing the facility support on the numerical calculations in this paper.

\newpage
\bibliography{references,custom}

\newpage

\appendix

\section{ICT Datasets}

\subsection{ICTQA}
\label{sec:appendix_ictqa_dataset}
To analyze various question types, we follow the classification method of \citet{yang-etal-2023-empower}. This approach categorizes questions based on their first interrogative word and assigns tags reflecting the nature of the information sought. The statistical data of this classification is detailed in Table~\ref{tab:ictqa_statistics}. Specifically, the ICTQA dataset labels questions under tags like `Parameter', `Configuration', and `Command', each indicating the type of information requested. For instance, `Parameter' relates to queries about specific values or settings, while `Configuration' pertains to questions regarding the setup of systems or processes. Additionally, questions are grouped by their first interrogative word. This categorization sheds light on user inquiries: `What' typically seeks factual details, and `How' focuses on procedures or techniques. The average length of questions and answers in ICTQA is 75.13 characters and 160.25 characters, respectively. Table~\ref{tab:ictqa_examples} shows examples from ICTQA questions.

\begin{table}[h]
\centering\resizebox{0.46\textwidth}{!}
{
\begin{tabular}{ll|lc} 
\toprule
\multicolumn{2}{c|}{\textbf{Question Tag (\%)}}        & \multicolumn{2}{c}{\textbf{$1^{st}$ Question word (\%)}} \\ \hline

Parameter               & 19.55                 & What                     & 29.33             \\
Configuration           & 17.94                 & How                      & 16.84                   \\
Command                 & 12.25                 & Why                      & 11.6                \\
Other                   & 50.26                 & Which                    & 9.14                  \\ \cline{1-2}
\multicolumn{2}{c|}{\textbf{Avg \# of length}} & Can                        & 6.57                     \\ \cline{1-2}
Question                & 75.13              & Is                        & 4.24                     \\
Answer                  & 160.25              & Other                    & 22.28                   \\ \bottomrule
\end{tabular}
\vspace{-3mm} 
}
\caption{Statistics of ICTQA}
\label{tab:ictqa_statistics}
\vspace{-4mm}
\end{table}


\begin{table}[t]
\small
\begin{tcolorbox}

Question 1:
\\
What is the range for the "number" parameter when configuring the maximum number of routes supported by the VPN instance IPv4 address family on the USG9500?\\
\\
\textcolor{blue}{Answer 1:\\The value for the "number" parameter can range from 1 to 500,000 on the USG9500.}\\
\rule{\linewidth}{0.2mm}
Question 2:
\\
Can NetStream sampling be enabled on the ingress or transit node for traffic over Segment Routing tunnels?\\
\\
\textcolor{blue}{Answer 2:\\Sampling on the ingress or transit node is not supported.}\\
\rule{\linewidth}{0.2mm}
Question 3:
\\
How can I monitor the connectivity between a MEP and an RMEP or between a MEP and a MIP on other devices using 802.1ag MAC ping?\\
\\
\textcolor{blue}{Answer 3:\\Run ping mac-8021ag mep mep-id mep-id [ md md-name ma ma-name ] { mac mac-address | remote-mep mep-id mep-id } [ -c count | -s packetsize | -t timeout | -p priority-value ].}\\

\end{tcolorbox}
\caption{Three examples from the ICTQA dataset.}
\label{tab:ictqa_examples}
\end{table}

\begin{table}[]
\centering\resizebox{0.43\textwidth}{!}
{
\begin{tabular}{l|c}
\toprule
\textbf{Statistics}   & \textbf{Value}  \\ \hline
Total Word Count in ICT-DATA     & 987M  \\
Total Word Count in ICT Tables  & 178M    \\
Average Word Count per Table    & 476.71  \\
Average Number of Cells per Table    & 13.29  \\
Average Text Length per Cell  & 35.87  \\
\bottomrule
\end{tabular} 
} 
\caption{ICT-DATA Statistical Overview}
\label{tab:table_cell_Statistics}
\end{table}

\begin{figure}[b]
\centering 
     \includegraphics[width=0.48\textwidth]{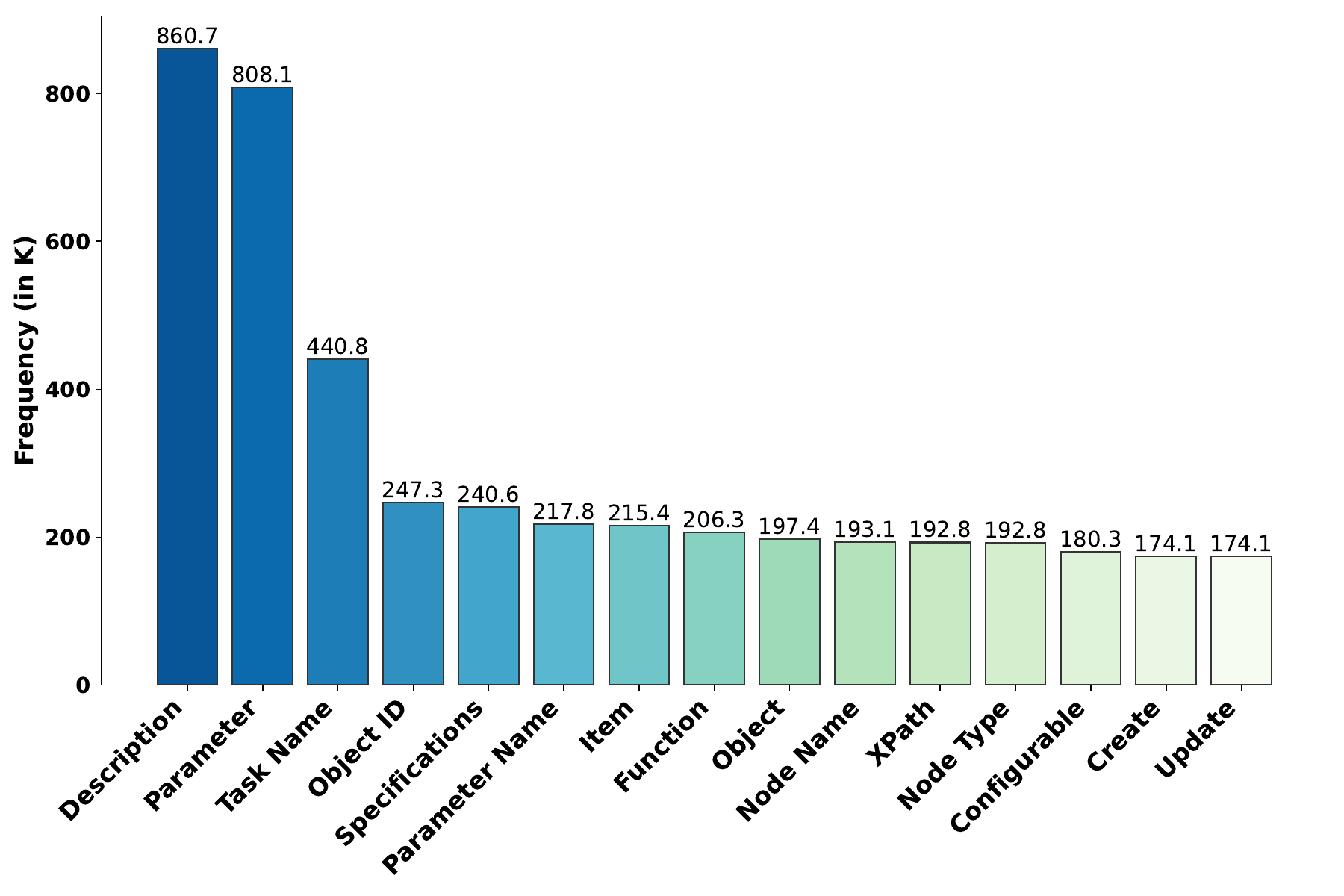}
     \caption{Top 15 Frequent Cell Contents in the Header Row of Tables.}
     \label{fig:top_15_header_row}
\end{figure}

\begin{table}[b]
\small
\begin{tcolorbox}

Below is an instruction that describes a task, paired with an input that provides further context. Write a response that appropriately completes the request.

\textbf{Instruction:}
\textcolor{blue}{Please answer the following questions concerning ICT products.}

\textbf{Input:}
\textcolor{blue}{\{Question\}}

\textbf{Response:}
\textcolor{blue}{\{Answer\}}

\end{tcolorbox}
\vspace{-2mm}
\caption{The instruction template.}
\label{tab:instruction_examples}
\end{table}

\subsection{ICT-DATA}
\label{sec:appendix_ictdata_dataset}
In the process of collecting the ICT-DATA, we perform preprocessing on the table data. Specifically, to standardize the tables from the dataset, we transform them into N x M arrays. For tables with merged cells, we expand the col-span or row-span attributes, copying the content into individual cells. Additionally, to illustrate the characteristics of tables in the ICT domain, Figure~\ref{fig:top_15_header_row} shows the top 15 frequent cell contents in the header rows of all tables in the ICT-DATA dataset. Table \ref{tab:table_cell_Statistics} provides a detailed statistical overview of the ICT-DATA. The total number of words in the dataset reaches 987 million, of which there are 178 million words in the tables, accounting for about 18\% of the total dataset. On average, each table contains about 477 words, about 13 cells, and the average text length of each cell is about 36 words.

\subsection{Instruction Template}
\label{sec:appendix_instruction_template}

Table~\ref{tab:instruction_examples} shows the instruction template we use. We fill the question and answer slots in the template with the QA pairs from the ICTQA dataset to form a set of instructions.

\section{Table-to-Text Generation Setups}
\label{sec:appendix_table_to_text_details_examples}

\subsection{Template Design for Table Serialization}

The tables in the ICT-DATA dataset consist of two types: relational tables and key-value pair tables. These two types of tables can be easily distinguished by matching keywords in the header row cells and considering the number of columns in the table. As illustrated in Table~\ref{tab:template_examples}, we develop distinct templates for each type of table: 1) Key-value pair tables, as shown in Table~\ref{tab:table_to_text_examples_1}, contain $m$ key-value pairs that describe entities mentioned in the table's title. 2) Relational tables, as shown in Table~\ref{tab:table_to_text_examples_2}, include a main column (MC) and $n$ attribute columns (AC), where the cells in the main column represent the entities described by the table. The main column can be identified through simple rules, including the uniqueness of its content and the presence of specific keywords in the header row.
To enhance the diversity of the text produced through the Template serialization, we compile a specialized glossary. When a term from this glossary is found in a table's header, the corresponding template content is adjusted accordingly. For example, if the string in the main column is ``Name'', ``The [$AC_1$] of the [$MC$] named [$C_{M_1}$] is [$C_{M_1A_1}$].'' will be changed as ``The [$AC_1$] of the [$C_{M_1}$] is [$C_{M_1A_1}$].''

\begin{table}[]
\centering
\small
\begin{tcolorbox}

\textbf{Template 1 for relational tables}\\
\vspace{-0.1cm}

\resizebox{1\textwidth}{!}
{
\begin{tabular}{|c|c|c|c|c|} 
\multicolumn{5}{c}{Table Title} \\ \hline
\multicolumn{1}{|c|}{MC} & \multicolumn{1}{c|}{$AC_1$} & \multicolumn{1}{c|}{$AC_2$}  & \ldots & \multicolumn{1}{c|}{$AC_n$}  \\ \hline
$C_{M_1}$ & $C_{M_1A_1}$ & $C_{M_1A_2}$ & \ldots & $C_{M_1A_n}$  \\ \hline
\multicolumn{5}{|c|}{\ldots} \\ \hline
$C_{M_m}$ & $C_{M_mA_1}$ & $C_{M_mA_2}$ & \ldots & $C_{M_mA_n}$ \\  \hline 
\end{tabular}
}

\vspace{0.3cm}

\textcolor{blue}{Generated Text:\\
The following sentences describe [Table Title].
The [$AC_1$] of the [$MC$] named [$C_{M_1}$] is [$C_{M_1A_1}$].
Its [$AC_2$] is [$C_{M_1A_2}$].
\ldots
Its [$AC_n$] is [$C_{M_1A_{n-1}}$].
\ldots
The [$AC_1$] of the [$MC$] named [$C_{M_m}$] is [$C_{M_mA_1}$].
\ldots
Its [$AC_n$] is [$C_{M_mA_{n-1}}$].
}

\rule{\linewidth}{0.15mm} \\

\textbf{Template 2 for key-value pair tables}\\
\vspace{-0.1cm}
\begin{center}
\resizebox{0.55\textwidth}{!}
{
\begin{tabular}{|c|c|} 
\multicolumn{2}{c}{Table Title} \\ \hline
\multicolumn{1}{|c|}{Header 1} & \multicolumn{1}{c|}{Header 2} \\ \hline

$K_1$ & $V_1$  \\ \hline
$K_2$ & $V_2$  \\ \hline
\multicolumn{2}{|c|}{\ldots} \\ \hline
$K_m$ & $V_m$  \\   \hline 

\end{tabular}
}
\end{center}
\vspace{0.15cm}

\textcolor{blue}{Generated Text:\\
The following sentences describe [Table Title].
The [$K_1$] of [$C_P$] is [$V_1$].
Its [$K_2$] is [$V_2$].
\ldots
Its [$K_m$] is [$V_m$].
}

\end{tcolorbox}
\caption{Two Templates in the Template Method.}
\label{tab:template_examples}
\end{table}

\subsection{Prompt for LLM-based Method}
In Table~\ref{tab:prompt_template_llm_based_t2t}, we present a prompt template specifically designed for the LLM-based table-to-text method. This template is tailored for generating natural language descriptions from tables in a two-dimensional array format, and includes a demonstration. In the prompt, we instruct LLMs not to output any additional information (which does not appear in the table, but comes from the internal knowledge of the LLM), regardless of whether this information is relevant to the table content.

\begin{table}[h]
\small
\begin{tcolorbox}
Now you have a task to complete. Task description: You will be given a table (with the 2d array format with the Caption). You need to generate a natural language description of the contents of the table. You can only generate content from the table content, do not generate other related or unrelated content. Here is an examples.\\

\textbf{Table:} \textcolor{brown}{
Caption: Parameters for the ip link add name and ip link del dev.
\\ {[['Parameter', 'Description', 'Value'], ['name NAME', 'Specifies the name of a bridge.', 'The value is a string of 1 to 15 case-sensitive characters without spaces.'], ['dev DEV', 'Specifies the name of a bridge.', 'The value is a string of 1 to 15 case-sensitive characters without spaces.'], ['type bridge', 'Indicates that the device type is bridge.', '-']].}}

\textbf{Description:} \textcolor{brown}{The table provides details on the parameters for the ip link add name and ip link del dev commands. There are different parameters for configuring a bridge. The "name NAME" parameter is for specifying the name of a bridge and accepts a string with 1 to 15 case-sensitive characters, excluding spaces. The "dev DEV" parameter also specifies the name of a bridge and requires a string of 1 to 15 case-sensitive characters without spaces. The "type bridge" parameter indicates that the device being configured is of the type 'bridge.' It does not require a specific value.}\\

\textbf{Table:} \textcolor{blue}{\{Table\}}

\textbf{Description:}

\end{tcolorbox}
\vspace{-2mm}
\caption{The prompt template designed for the LLM-based table-to-text generation.}
\label{tab:prompt_template_llm_based_t2t}
\end{table}

\subsection{Table-to-Text Generation Examples}
In Table~\ref{tab:table_to_text_examples_1}, we showcase the conversion of a simple two-column table into text using four different methods. For a more complex scenario involving a multi-column table with empty cells, refer to the example provided in Table~\ref{tab:table_to_text_examples_2}. Lastly, the adaptation of these methods for a table featuring both multiple columns and merged cells is displayed in Table~\ref{tab:table_to_text_examples_3}.

\begin{table*}[t]
\centering\resizebox{0.7\textwidth}{!}{
\begin{tabular}{lccccc}
\toprule
\multirow{2}{*}{\textbf{Model}}  & \textbf{Pre-training Cost} & \textbf{Fine-Tuning Cost} & \textbf{Total Cost} & \multirow{2}{*}{\textbf{GPU Node}} \\ 
& (GPU hours) & (GPU hours) & (GPU hours) & \\ \hline
OPT-1.3B     & 102   & 1      & 103  & 8*V100-32G \\
OPT-2.7B     & 176   & 2      & 178      & 8*V100-32G    \\
OPT-6.7B     & 339   & 5      & 344     & 8*V100-32G    \\
OPT-13B      & 660   & 5      & 665     & 8*V100-32G     \\
Llama2-7B    & 519   & 5      & 524     & 8*V100-32G     \\
Llama2-13B   & 549   & 7      & 556     & 4*A100-40G     \\

\bottomrule
\end{tabular}
}
\vspace{-2mm}
\caption{The GPU cost of training each model under the QLoRA strategy using a corpus generated by the \textbf{Markdown} table-to-text generation method. The training cost with corpora produced by other methods is close to that of Markdown. The GPU hours are computed as follows: (iteration time (seconds) $\times$ number of iterations $\times$ number of GPUs ÷ 3600 seconds/hour).}
\label{tab:gpu_costs}
\end{table*}

\section{Training Setup and GPU costs}
\label{sec:appendix_training_setup}

This paper involves model training within the DSFT QA framework. We utilize an A100 40GB node equipped with 4 GPUs for both pre-training and fine-tuning of the Llama2-13B model. The pre-training and fine-tuning phases for other models are performed on a V100 32GB node with 8 GPUs. These processes leverage the DeepSpeed framework \cite{rasley2020deepspeed}. Table~\ref{tab:gpu_costs} provides a detailed overview of the GPU costs for training each model under the markdown method setting.

During the pre-training stage, we adopt an unsupervised learning approach, focusing on next-token prediction. To optimize memory usage, the DeepSpeed Zero Redundancy Optimizer (Zero Stage 2) is employed \cite{rajbhandari2020zero}. Training parameters include a per-GPU batch size of 16 and a fixed seed value (1234) to ensure reproducibility. Each model undergoes a single epoch of training, utilizing a cosine learning rate scheduler with an initial rate of 2e-4. The learning rate warmup ratio is set at 0.05, accompanied by a weight decay rate of 0.01. Data preprocessing is performed with a block size of 512. In the QLoRA configuration, trainable parameters include the transformer's query, key, value, and output projection matrices, along with token embeddings and the language model head. Other parameters include: a LoRA rank of 64, an alpha value of 128, a dropout rate of 0.05 for the LoRA layer, and float16 for PyTorch \cite{paszke2019pytorch} tensors.

For the instruction fine-tuning phase, we derive the instruction dataset from the ICTQA training set. The training parameters for this phase are: 5 training epochs, a maximum token length of 512, a batch size of 8 per GPU, and a learning rate of 1e-4 with a cosine decaying scheduler. The configuration of QLoRA remains the same as in the pre-training phase.

\section{Evaluation Setup}
\label{sec:appendix_evaluation_setup}

\subsection{Prompt for LLM as Evaluator}

\begin{table*}[]
\small
\begin{tcolorbox}

[System]\\
You will be provided with a question, the correct answer to the question, and four candidate answers. Your responsibility is to evaluate the consistency between the candidate answers and the correct answer. The focus should be on understanding the correlation or similarity of the content, rather than grammar or style. Please make sure you understand these guidelines before proceeding. \\

\textcolor{blue}{Consult this guide whenever needed:\\
0, penalty:\\
The candidate answers have issues such as repetitive sentences, which can significantly impair the helpfulness of the response.\\
1, Very low correlation:\\
Indicates that the candidate answer is almost entirely unrelated or opposite to the correct answer.\\
2, Low correlation:\\
Indicates that the candidate answer significantly deviates from the correct answer.\\
3, Moderate correlation:\\
Suggests that the candidate answer shares some similarities with the correct answer but may lack several key points or include extra unrelated content.\\
4, High correlation:\\
Indicates that the candidate answer is largely consistent with the correct answer, missing only minor points or details.\\
5, Very high correlation:\\
Signifies that the candidate answer is almost identical to or captures the complete essence of the correct answer.}\\

You will need to categorize the four candidate answers A, B, C, and D based on their relevance.

For example, A:1, B:2, C:3, D:5 means: A's correlation with the correct answer falls into 1. Very low correlation. B's correlation with the correct answer is higher than A's, at 2. Low correlation. C's correlation with the correct answer is 3. Moderate correlation. D's correlation with the correct answer is 5. Very high correlation.\\

\textcolor{red}{<A demonstration of the evaluation>}\\

[User]\\
Evaluation Form (only score, do not output any other explanation):\\
<Question>: \textcolor{blue}{question}\\
<Correct answer> (Standard Answer): \textcolor{blue}{golden answer}\\
A answer: \textcolor{brown}{response\_1}\\
B answer: \textcolor{cyan}{response\_2}\\
C answer: \textcolor{magenta}{response\_3}\\
D answer: \textcolor{orange}{response\_4}\\
Score:

\end{tcolorbox}
\caption{The prompt of the LLM evaluator gives scores on four response candidates.}
\label{tab:prompt_template_for_GPT4_evaluation}
\end{table*}

\begin{table*}[]
\small
\begin{tcolorbox}

<Question>: \textcolor{blue}{How long can the Information field be in a PPP data packet?}\\
<Correct answer> (Standard Answer): \textcolor{blue}{The maximum length for the Information field, including the Padding field, is the maximum receive unit (MRU). The MRU defaults to 1500 bytes and can be negotiated.}\\

A answer: \textcolor{brown}{The length of the information field is limited to 1500 bytes.}

B answer: \textcolor{cyan}{The information field can be up to 1500 bytes in length. The maximum size of the information field is specified by RFC 1661 and is set at 1500 bytes. This allows for the transmission of large packets, but also ensures that the protocol remains efficient and reliable.}

C answer: \textcolor{magenta}{The length of the information field is variable and can range from 0 to 1536 bytes.}

D answer: \textcolor{orange}{The length of the information field is not specified, but it must be at least 1 byte. The information field is variable length and contains the protocol specific information that the peer requires to establish the link. The length of the information field is not specified, but it must be at least one byte. The format of the information field is defined by the protocol being used. For example, IPX uses the information field to specify the IPX network number and the IPX node address.}\\

Score: \textcolor{violet}{A:5, B:5, C:4, D:2}
 
\end{tcolorbox}
\caption{One-Shot Demonstration for the LLM Evaluator.}
\label{tab:example_in_gpt4_eval}
\end{table*}

Table~\ref{tab:prompt_template_for_GPT4_evaluation} details the prompt template used for evaluating responses via LLM (GPT-4) in a one-shot setting. The task of the GPT-4 is to compare four responses to a grounded answer and assigning each a score ranging from 0 to 5. Additionally, Table~\ref{tab:example_in_gpt4_eval} provides an example of the one-shot demonstration used in this evaluation.

\subsection{Setup and Criteria for Human Evaluation}

\begin{table*}[]
\small
\begin{tcolorbox}

The scoring criteria for human evaluation:\\

0, penalty:\\
The candidate answers have issues such as repetitive sentences, which can significantly impair the helpfulness of the response.\\

1, Very low correlation:\\
Indicates that the candidate answer is almost entirely unrelated or opposite to the correct answer.\\

2, Low correlation:\\
Indicates that the candidate answer significantly deviates from the correct answer.\\

3, Moderate correlation:\\
Suggests that the candidate answer shares some similarities with the correct answer but may lack several key points or include extra unrelated content.\\

4, High correlation:\\
Indicates that the candidate answer is largely consistent with the correct answer, missing only minor points or details.\\

5, Very high correlation:\\
Signifies that the candidate answer is almost identical to or captures the complete essence of the correct answer.\\

\end{tcolorbox}
\caption{The scoring criteria for human evaluation.}
\label{tab:scoring_criteria_for_human_eval}
\end{table*}

For the human evaluation, three co-authors of this paper, all with domain expertise in ICT products, are designated as evaluators. Each sample receives three independent evaluations from these qualified evaluators. We analyze the consistency of scoring across the three evaluations. If the ranking order of the four responses remains consistent and the score difference for the same response across different evaluators does not exceed one point, the evaluation is deemed reliable. In cases of significant discrepancies, evaluators are requested to reassess the sample. In all evaluation documents, the sources of the responses are anonymously presented as `A',`B',`C', and `D'. Table~\ref{tab:scoring_criteria_for_human_eval} shows the scoring criteria for human evaluation, which is consistent with the scoring criteria for LLM as evaluator presented in Table~\ref{tab:prompt_template_for_GPT4_evaluation}.

\begin{table*}[]
\small
\begin{tcolorbox}

Table Example 1: Basic information about the PLCh-Power-1
\begin{center}
\includegraphics[width=0.97\textwidth]{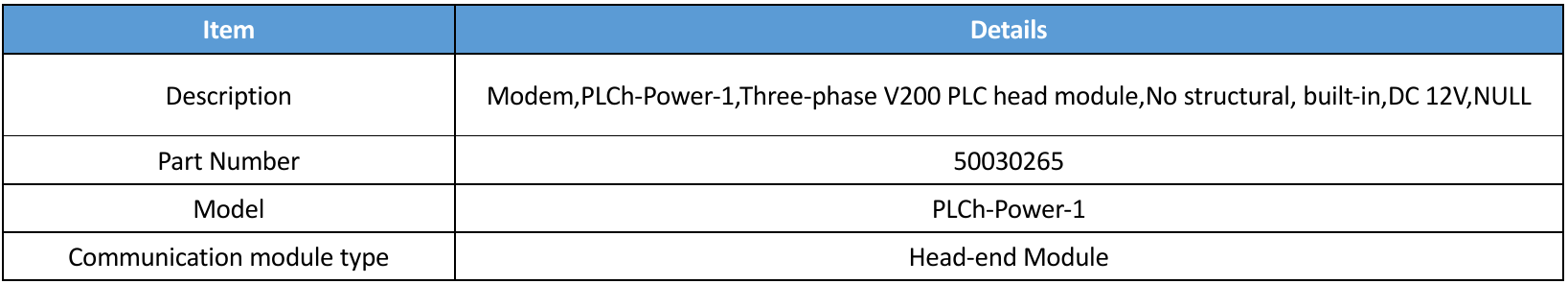}
\end{center}
Markdown:
\textcolor{purple}{\\
Table Caption: Basic information about the PLCh-Power-1 \\
| Item | Details |\\
| :--- | :--- |\\
| Description | Modem,PLCh-Power-1,Three-phase V200 PLC  head module,No structural, built-in,DC 12V,NULL | \\
| Part Number | 50030265 | \\
| Model | PLCh-Power-1 | \\
| Communication module type | Head-end Module | }\\

Template:
\textcolor{black}{
The following sentences describe about Basic information about the PLCh-Power-1. 
The Description of the PLCh-Power-1 is ``Modem,PLCh-Power-1,Three-phase V200 PLC head module,No structural, built-in,DC 12V,NULL''. 
Its Part Number is 50030265. 
Its Model is PLCh-Power-1. 
Its Communication module type is Head-end Module. }
\\

TPLM-based:
\textcolor{magenta}{The description item is Modem,PLCh-Power-1,Three-phase V200 PLC  head module,No structural, built-in,DC 12V,NULL. The part number item is 50030265. The model item is PLCh-Power-1. The communication module type item is Head-end Module.}
\\

LLM-based:
\textcolor{blue}{The table provides details about the description, part number, model, and communication module type. The item described is a Modem, specifically the PLCh-Power-1, which is a three-phase V200 PLC head module. It is noted that it has no structural built-in and operates on DC 12V. The part number for this item is 50030265, and the communication module type is identified as a Head-end Module.}\\

\end{tcolorbox}
\caption{A table-to-text example of four methods for a simple table with only two columns.}
\label{tab:table_to_text_examples_1}
\end{table*}

\begin{table*}[]
\small
\begin{tcolorbox}

Table Example 2: Indicators on the PLC-IH-1
\begin{center}
\includegraphics[width=0.97\textwidth]{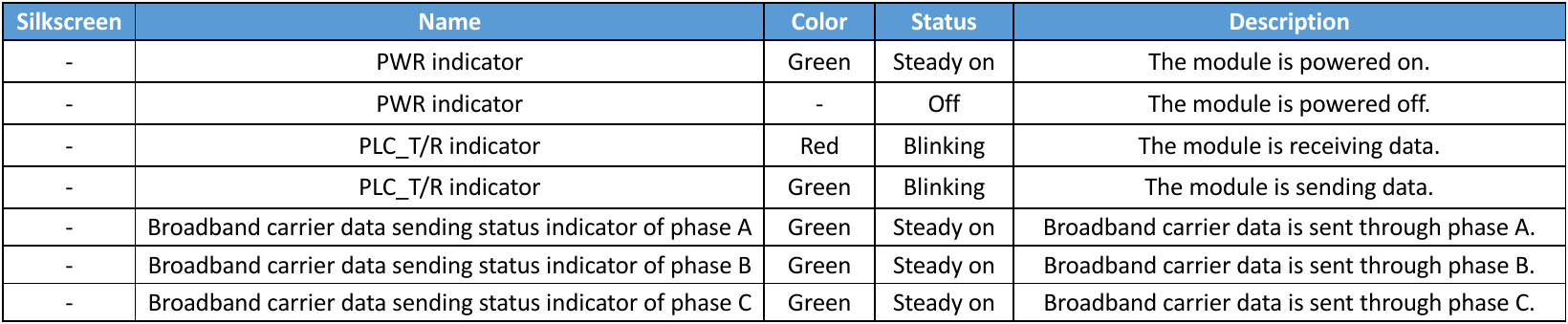}
\end{center}
Markdown:\\
\textcolor{purple}{
Table Caption: Indicators on the PLC-IH-1 \\
|Silkscreen| Name | Color| Status| Description|\\
|:---|:---|:---|:---|:---|\\
|-| PWR indicator| Green| Steady on| The module is powered on.|\\
| - | PWR indicator | - | Off| The module is powered off.|\\
| - | PLC\textunderscore{}T/R indicator| Red| Blinking  | The module is receiving data.|\\
| - | PLC\textunderscore{}T/R indicator | Green   | Blinking  | The module is sending data.|\\
| - | Broadband carrier data sending status indicator of phase A | Green|Steady on | Broadband carrier data is sent through phase A. |\\
| -            | Broadband carrier data sending status indicator of phase B | Green   | Steady on | Broadband carrier data is sent through phase B. |\\| -            | Broadband carrier data sending status indicator of phase C | Green   | Steady on | Broadband carrier data is sent through phase C. |
}\\

Template:
\textcolor{black}{The following sentences describe about Indicators on the PLC-IH-1. The Color of the PWR indicator is Green. Its Status is Steady on. Its Description is The module is powered on. The Status of the PWR indicator is Off. Its Description is The module is powered off. The Color of the PLC\_T/R indicator is Red. Its Status is Blinking. Its Description is The module is receiving data. The Color of the PLC\_T/R indicator is Green. Its Status is Blinking. Its Description is The module is sending data. The Color of the Broadband carrier data sending status indicator of phase A is Green. Its Status is Steady on. Its Description is Broadband carrier data is sent through phase A. The Color of the Broadband carrier data sending status indicator of phase B is Green. Its Status is Steady on. Its Description is Broadband carrier data is sent through phase B. The Color of the Broadband carrier data sending status indicator of phase C is Green. Its Status is Steady on. Its Description is Broadband carrier data is sent through phase C.  }
\\

TPLM-based:
\textcolor{magenta}{The - silkscreen ( also known as PWR indicator ) is a function that can be expressed as a function. The - silkscreen is green. The - silkscreen is Steady on. The - silkscreen has the symbol The module is powered on.. The - silkscreen ( also known as PWR indicator ) is a function that can be expressed as a function. The - silkscreen is -. The - silkscreen is off. The - silkscreen has the symbol The module is powered off.. The - silkscreen ( also known as PLC\textunderscore{}T/R indicator ) is a function that can be expressed as a function. The - silkscreen is red. The - silkscreen is blinking. The - silkscreen has the symbol The module is receiving data.. The - silkscreen ( also known as PLC\textunderscore{}T/R indicator ) is a function that can be expressed as a function. The - silkscreen is green. The - silkscreen is blinking. The - silkscreen has the symbol The module is sending data.. The - silkscreen ( also known as Broadband carrier data sending status indicator of phase A ) is a function that can be expressed as a function. The - silkscreen is green. The - silkscreen is Steady on. The - silkscreen has the symbol Broadband carrier data is sent through phase A.. The - silkscreen ( also known as Broadband carrier data sending status indicator of phase B ) is a function that can be expressed as a function. The - silkscreen is green. The - silkscreen is Steady on. The - silkscreen has the symbol Broadband carrier data is sent through phase B.. The - silkscreen ( also known as Broadband carrier data sending status indicator of phase C ) is a function that can be expressed as a function. The - silkscreen is green. The - silkscreen is Steady on. The - silkscreen has the symbol Broadband carrier data is sent through phase C..}
\\

LLM-based:
\textcolor{blue}{The table provides information about various indicators on a module. The "Silkscreen" column specifies the name of each indicator, while the "Color" column indicates the color of the indicator. The "Status" column describes the behavior of the indicator, such as whether it is steady on or blinking. The "Description" column provides additional details about the indicator's function. 
There are two indicators named "PWR indicator" in the table. The first one is green in color and is steady on, indicating that the module is powered on. The second "PWR indicator" is not specified for color and is off, indicating that the module is powered off. 
There are two indicators named "PLC\textunderscore{}T/R indicator" in the table. The first one is red in color and is blinking, indicating that the module is receiving data. The second "PLC\textunderscore{}T/R indicator" is green in color and is also blinking, indicating that the module is sending data. 
There are three indicators related to broadband carrier data sending status, each associated with a specific phase (A, B, and C). All three indicators are green in color and are steady on, indicating that broadband carrier data is being sent through each respective phase.
}\\

\end{tcolorbox}
\caption{A table-to-text example of four methods for a table with multiple columns and empty cells.}
\label{tab:table_to_text_examples_2}
\end{table*}

\begin{table*}[]
\small
\begin{tcolorbox}

Table Example 3: Relationship between the device group type and device networking reliability
\begin{center}
\includegraphics[width=0.97\textwidth]{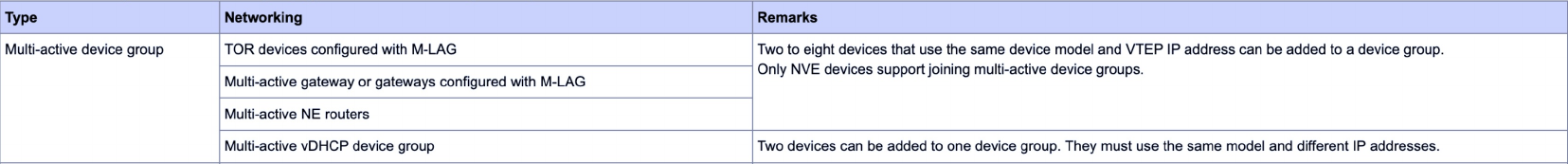}
\end{center}
Markdown:\\
\textcolor{purple}{
Table Caption: Relationship between the device group type and device networking reliability \\
| Type | Networking | Remarks |\\
 | :--- | :--- | :--- | \\
 | Multiple-active device group | ToR devices configured with M-LAG | Two to eight devices that use the same device model and VTEP IP address can be added to a device group. | \\
 | Multiple-active device group | Multiple-active gateway or gateways configured with M-LAG | Two to eight devices that use the same device model and VTEP IP address can be added to a device group. | \\
 | Multiple-active device group | Multiple-active NE routers | Two to eight devices that use the same device model and VTEP IP address can be added to a device group. | \\
 | Multiple-active device group | Multiple-active vDHCP device group | Two to eight devices that use the same device model and VTEP IP address can be added to a device group. |
}\\

Template:
\textcolor{black}{The following sentences describe about Relationship between the device group type and device networking reliability. The Networking of the Type named Multiple-active device group is ToR devices configured with M-LAG. Its Remarks is Two to eight devices that use the same device model and VTEP IP address can be added to a device group. The Networking of the Type named Multiple-active device group is Multiple-active gateway or gateways configured with M-LAG. Its Remarks is Two to eight devices that use the same device model and VTEP IP address can be added to a device group. The Networking of the Type named Multiple-active device group is Multiple-active NE routers. Its Remarks is Two to eight devices that use the same device model and VTEP IP address can be added to a device group. The Networking of the Type named Multiple-active device group is Multiple-active vDHCP device group. Its Remarks is Two to eight devices that use the same device model and VTEP IP address can be added to a device group. }
\\

TPLM-based:
\textcolor{magenta}{The multiple-active device group type is ToR devices configured with M-LAG. The multiple-active device group type is Two to eight devices that use the same device model and VTEP IP address can be added to a device group.. The multiple-active device group type is Multiple-active gateway or gateways configured with M-LAG. The multiple-active device group type is Two to eight devices that use the same device model and VTEP IP address can be added to a device group.. The multiple-active device group type is Multiple-active NE routers. The multiple-active device group type is Two to eight devices that use the same device model and VTEP IP address can be added to a device group.. The multiple-active device group type is Multiple-active vDHCP device group. The multiple-active device group type is Two to eight devices that use the same device model and VTEP IP address can be added to a device group.}
\\

LLM-based:
\textcolor{blue}{The table provides information about different types of device groups and their configurations. The "Multiple-active device group" type can be configured with various networking options such as ToR devices with M-LAG, multiple-active gateway or gateways with M-LAG, multiple-active NE routers, and multiple-active vDHCP device group. Each of these types allows for the addition of two to eight devices that share the same device model and VTEP IP address. 
}\\

\end{tcolorbox}
\caption{A table-to-text example of four methods for a table with multiple columns and merged cells.}
\label{tab:table_to_text_examples_3}
\end{table*}

\begin{figure}[ht]
    \centering
    \begin{subfigure}[t]{0.23\textwidth}
        \centering
        \includegraphics[width=\textwidth]{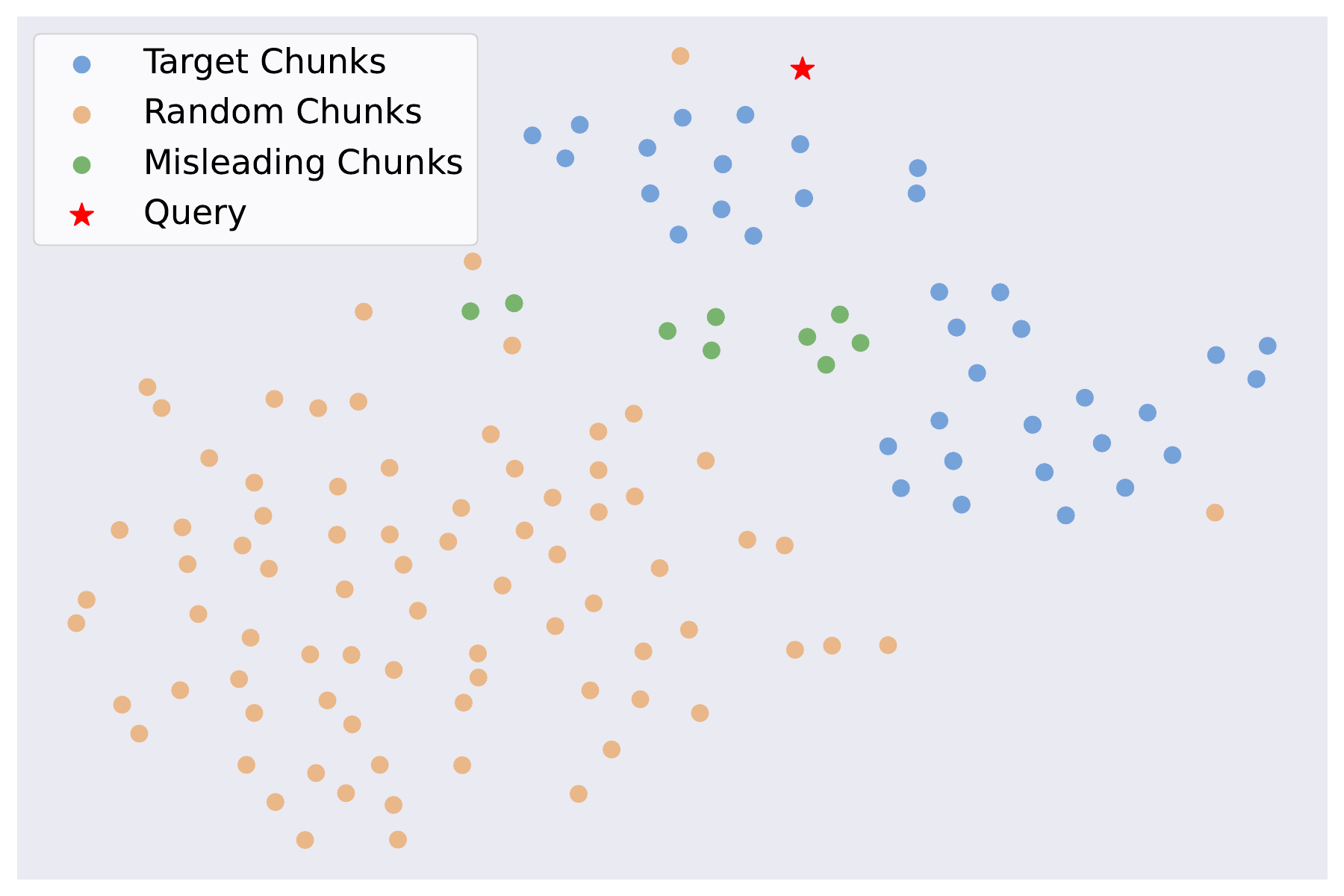}
        \caption{LLM-based}
    \end{subfigure}%
    \hfill
    \begin{subfigure}[t]{0.23\textwidth}
        \centering
        \includegraphics[width=\textwidth]{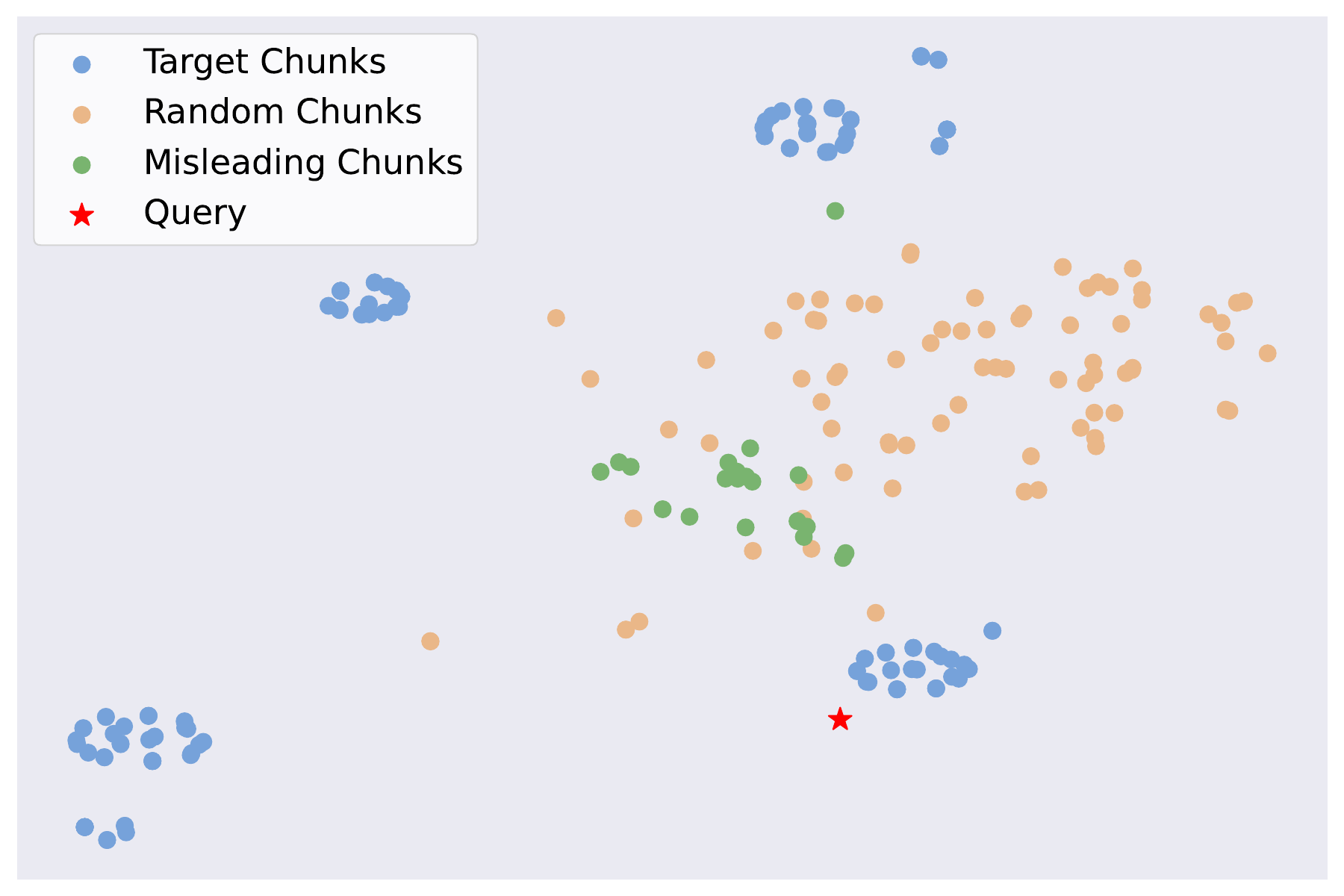}
        \caption{Markdown}
    \end{subfigure}
    

    \begin{subfigure}[b]{0.23\textwidth}
         \centering
         \includegraphics[width=\textwidth]{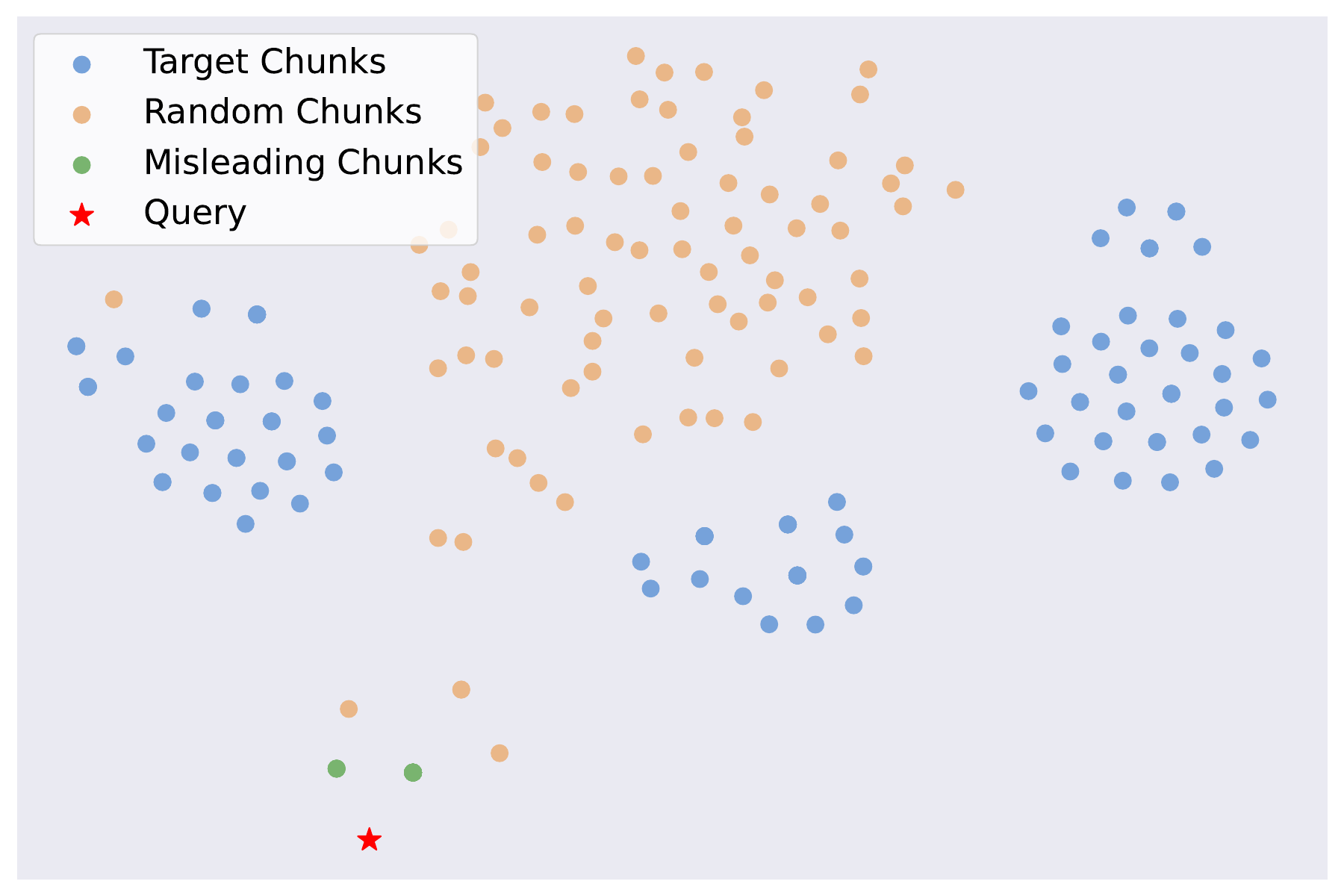}
         \caption{Template}
    \end{subfigure}
    \hfill
    \begin{subfigure}[b]{0.23\textwidth}
         \centering
         \includegraphics[width=\textwidth]{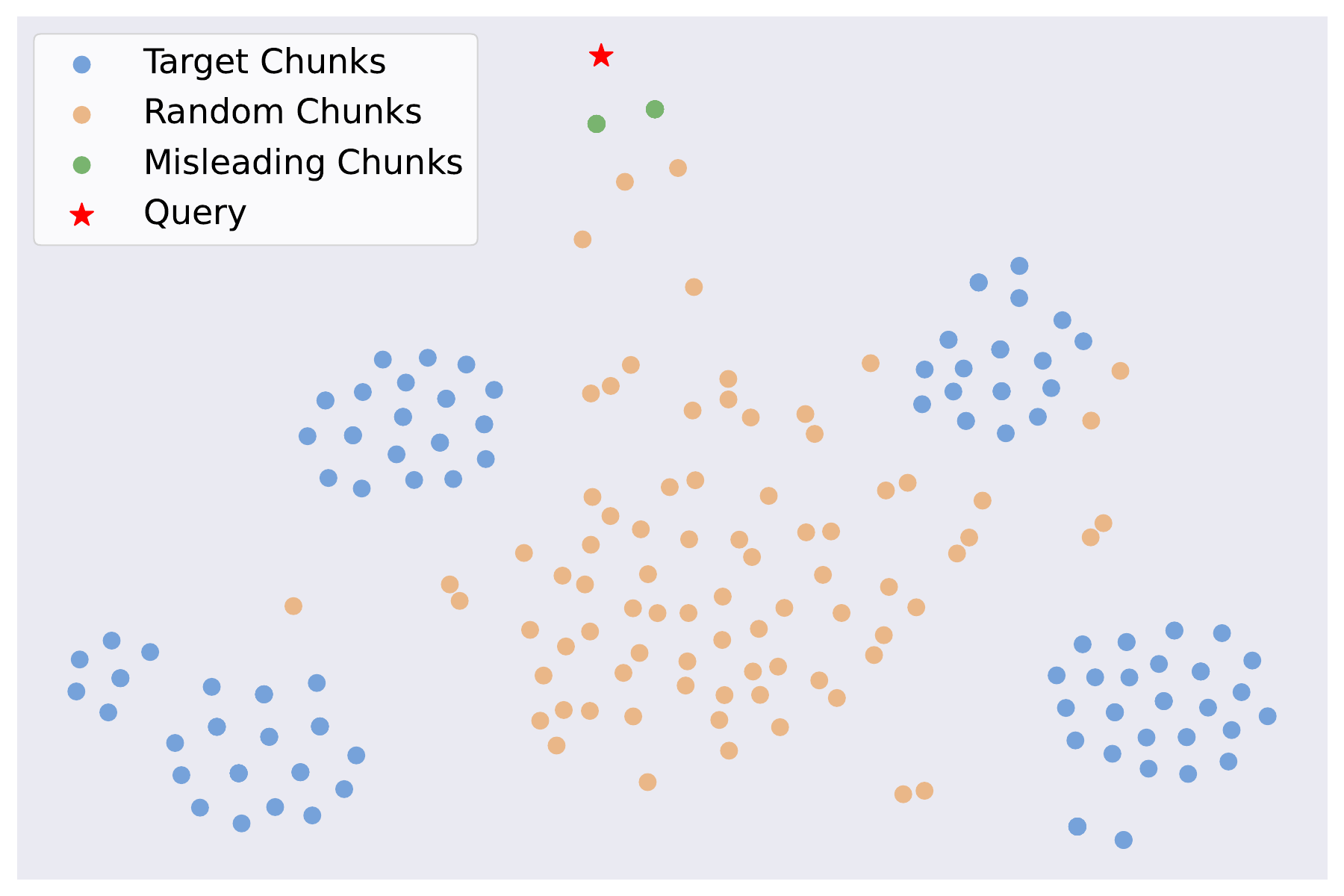}
         \caption{TPLM-based}
    \end{subfigure}
    
    \caption{T-SNE visualization of chunk clusters in the embedding space for the four table-to-text methods in the RAG system case study. `Random Chunks' represent chunks randomly selected from the corpus.}
    \label{fig:case_study_vis}
    \vspace{-4mm}
\end{figure}

\section{Case Analysis}
We demonstrate a QA case of a RAG QA system built upon the LLaMA-70B-chat model. As shown in Table \ref{llama70bllmcase}, our RAG QA system successfully retrieves the correct context information containing the query answer from the corpora generated by the Markdown and LLM-based methods. However, it fails to retrieve correct information from the corpora generated by the TPLM-based and Template methods. We show their t-SNE visualization in the semantic space in Figure~\ref{fig:case_study_vis}. In this case, it can be clearly seen that the misleading text chunks generated by the TPLM-based and template method, which are related to entities in the query but do not contain the correct answer, are semantically closer to the query. This leads to the failure to retrieve the correct chunks (i.e., Target chunks in the figure) containing the query’s answer, indicating that the text generated by these two methods has poor semantic representations.

\begin{table*}[]
\small
\begin{tcolorbox}

\textbf{Question}: How can I modify the entries in the pimBsrElectedBSRRPSetTable?
\\
\textbf{Gold Answer}: The entries in this table cannot be modified.
\\
\rule{\linewidth}{0.5mm}

\textbf{Markdown}\\
\textbf{Retrieved Passage}:  ... Title: \textcolor{red}{pimBsrElectedBSRRPSetTable} This table describes RP-Set information in the elected BSR ... Creation Restriction: No entry can be created in this table. Modification Restriction: \textcolor{red}{The entries in this table cannot be modified.} Deletion Restriction: The entries in this table cannot be deleted ...
\\
\textbf{Answer}: The modification restriction for the pimBsrElectedBSRRPSetTable is set to ``not-accessible,'' \textcolor{red}{indicating that the entries in this table cannot be modified.} This restriction is defined in the MIB file and cannot be changed through the command line or any other means. Therefore, it is not possible to modify the entries in this table. \textcolor{blue}{\checkmark}

\rule{\linewidth}{0.2mm}
\textbf{Template}\\
\textbf{Retrieved Passage}:  ... <Page\_Start>: Title: Configuration Precautions for VRRP <Page\_End> <Page\_Start>: Title: vrrpRouterStatsTable <Page\_End> <Page\_Start>: Title: Configuring VRRP ... 
\\
\textbf{Answer}: I don't know the answer. \textcolor{blue}{\ding{55}}

\rule{\linewidth}{0.2mm}
\textbf{TPLM-based}\\
\textbf{Retrieved Passage}: ... <Page\_Start>: Title: Configuration Precautions for VRRP <Page\_End> <Page\_Start>: Title: vrrpRouterStatsTable <Page\_End> <Page\_Start>: Title: Configuring VRRP ... 
\\
\textbf{Answer}: I don't know the answer. The knowledge provided does not contain any information about modifying entries in the pimBsrElectedBSRRPSetTable. The title of the knowledge suggests that it may be related to PIM (Protocol Independent Multicast) and BSR (Bootstrap Router), but without additional context or information, it is not possible to provide a helpful answer to the question.If you have any additional information or context, please provide it and I will do my best to assist you. \textcolor{blue}{\ding{55}}

\rule{\linewidth}{0.2mm}
\textbf{LLM-based}\\
\textbf{Retrieved Passage}: ... \textcolor{red}{pimBsrElectedBSRRPSetTable} This table describes RP-Set information in the elected BSR ... Creation Restriction: No entry can be created in this table. Modification Restriction: \textcolor{red}{The entries in this table cannot be modified}. Deletion Restriction: The entries in this table cannot be deleted...
\\
\textbf{Answer}: According to the provided knowledge, \textcolor{red}{it is not possible to modify the entries in the pimBsrElectedBSRRPSetTable.} The table provides information about various objects related to PIM BSR (Protocol Independent Multicast Bootstrap Router) elected BSR (Best Source Router) group mapping, and the entries in this table cannot be modified.The knowledge states that the creation restriction for this table is set to ``No entry can be created in this table,'' and the modification restriction is set to ``The entries in this table cannot be modified.'' Additionally, the deletion restriction is set to ``The entries in this table cannot be deleted.'' Therefore, it is not possible to modify the entries in the pimBsrElectedBSRRPSetTable, and any attempts to do so will likely result in an error message or a notification that the modification was unsuccessful. \textcolor{blue}{\checkmark}
\end{tcolorbox}
\caption{A QA example of the RAG QA system based on using a corpus generated by each of the four table-to-text methods as a retrieval source. The \textcolor{red}{red} font indicates text in the retrieved passage that is relevant to the answer.}
\label{llama70bllmcase}
\end{table*}

\end{document}